\documentclass[lettersize,journal]{IEEEtran}
\usepackage{amsmath,amsfonts}
\usepackage{algorithmic}
\usepackage{algorithm}
\usepackage{array}
\usepackage[caption=false,font=normalsize,labelfont=sf,textfont=sf]{subfig}
\usepackage{textcomp}
\usepackage{stfloats}
\usepackage{url}
\usepackage{verbatim}
\usepackage{graphicx}
\usepackage{multirow}
\usepackage{makecell}
\usepackage{amssymb}
\usepackage{cite}
\begin{document}
\bstctlcite{IEEEexample:BSTcontrol}

\title{MCRL2: Multi-resource Cross-attention-based Representation Learning-augmented Reinforcement Learning for Cloud Microservice Scheduling}

\author{Tiangang Li, Shi Ying, Xiangbo Tian, Chuan Shi, Ding Xiao
\thanks{Tiangang Li, Shi Ying and Xiangbo Tian are with the School of Computer Science, Wuhan University, Wuhan 430072, China. (Email: tiangangli@whu.edu.cn)}
\thanks{Chuan Shi and Ding Xiao are with the School of Computer Science, Beijing University of Posts and Telecommunications, Beijing 100876, China. }
}



\maketitle

\begin{abstract}
Efficient microservice scheduling is crucial for maintaining load balance across nodes in data centers and ensuring high quality of service. However, achieving this in practice remains challenging due to dynamic resource imbalance under fluctuating workloads, nonlinear coupling across multiple resource dimensions, and the heterogeneity of microservice resource demands. While reinforcement learning-based approaches have shown promise, they struggle to capture the complex interdependencies among heterogeneous resources and neglect the importance of learning informative system representations. To address these limitations, we propose MCRL2, a novel reinforcement learning approach augmented with multi-resource cross-attention-based representation learning for microservice scheduling. Specifically, we first propose MCRL, a novel representation learning approach that captures structured and informative interactions among nodes, resources, and microservices via a multi-resource cross-attention mechanism. Then, MCRL2 augments reinforcement learning through MCRL-enhanced actor-critic architecture combined with a maximum entropy objective, improving system state expressiveness and leading to more stable and effective scheduling decisions. Extensive experiments on real production cluster traces demonstrate that MCRL2 significantly outperforms existing baselines in load balancing, scheduling success rate and average completion time across diverse workload patterns.

\end{abstract}

\begin{IEEEkeywords}
Cloud computing, microservice scheduling, load balancing, representation learning, reinforcement learning,
\end{IEEEkeywords}

\section{Introduction}
\IEEEPARstart{D}{ue} to the inherent flexibility, microservices have become a widely adopted architecture for application development in cloud computing \cite{ref15} \cite{ref16} \cite{ref18} \cite{ref22} \cite{ref28}. With the rapid advancement of cloud computing, efficient microservice scheduling is crucial for maintaining the stable operation of cloud clusters \cite{ref13} \cite{ref19} \cite{ref20} \cite{ref29}. An analysis of the operational trajectories of 1,303 microservices in a large-scale production cloud cluster reveals several key challenges in microservice scheduling \cite{ref12}.

\begin{figure}[!t]
\centering
\includegraphics[width=0.45\textwidth]{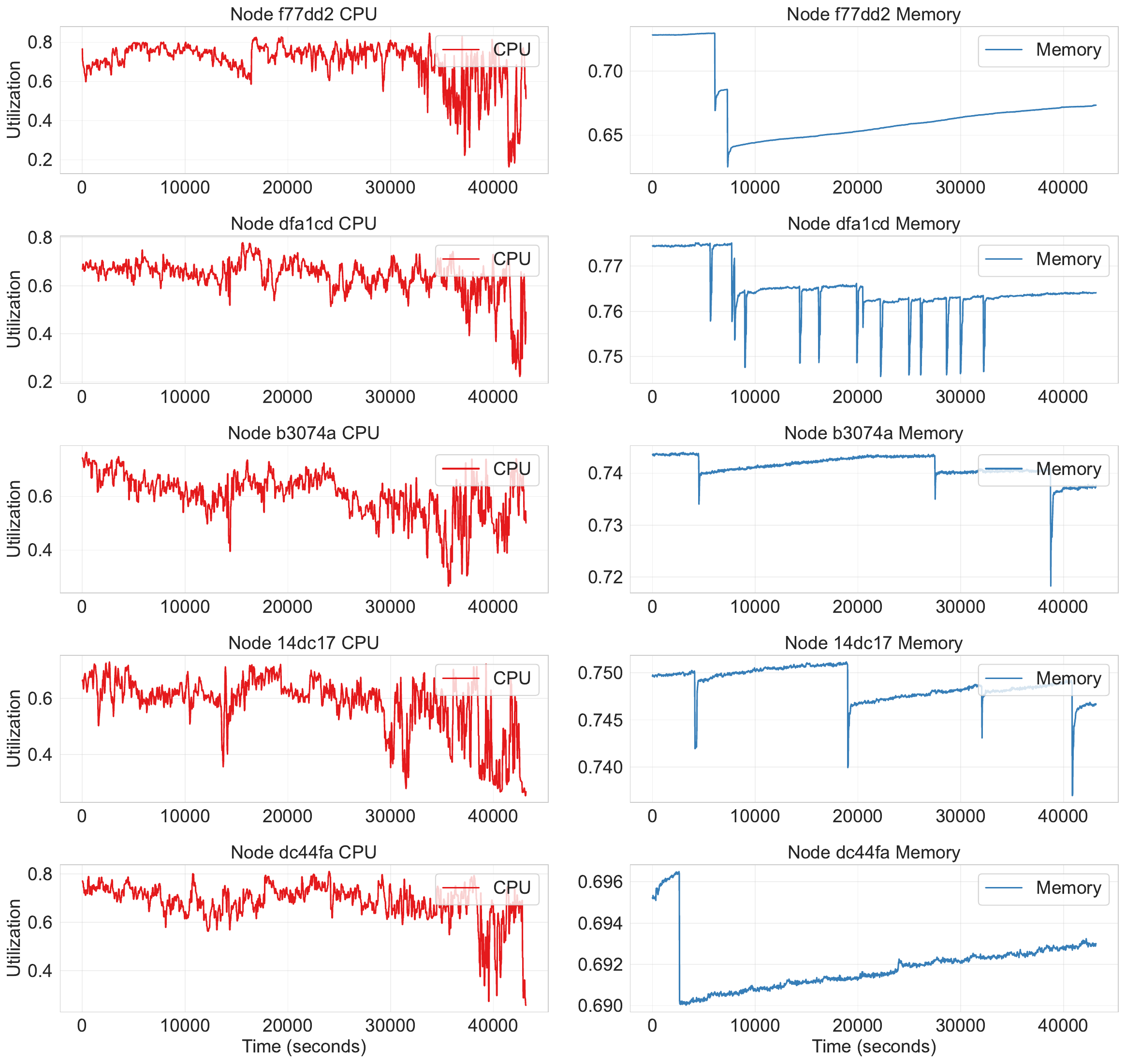}
\caption{CPU and memory utilization trends of 5 randomly selected nodes over time. The data is collected from a real-world Alibaba production cluster.}
\label{fig1}
\end{figure}

{\bf Dynamic resource imbalance across nodes under fluctuating workloads.} As illustrated in Fig. \ref{fig1}, the multi-dimensional resource utilization trends among nodes within the cluster exhibit significant differences over time, with mean distributions fluctuating in a non-stationary manner \cite{ref17}. Furthermore, the co-location of multiple microservice instances on the same node can exacerbate such imbalances, since resource contention caused by shared nodes often leads to short-term bursts or hotspots, amplifying the disparity in load distribution across nodes. This dynamic imbalance poses a challenge for schedulers, which must continually optimize node selection strategies under multi-dimensional, time-varying workload conditions to achieve stable global load balancing.

{\bf Dynamic nonlinear coupling effects across resource dimensions.}  As depicted in Fig. \ref{fig1}, observational data reveal complex, dynamic interdependencies among resource dimensions. The coupling between different resource dimensions is time-varying \cite{ref11}. At certain intervals or on specific nodes, negative correlations can be observed. For instance, periods of peak CPU utilization often correspond to lows in memory utilization. In contrast, other scenarios may exhibit positive or weak correlations across resource dimensions. This indicates that microservice workloads vary dynamically across different resource dimensions, leading to interdimensional resource contention. These patterns become more volatile under shared-node scenarios, where co-located instances introduce additional interference that intensifies nonlinear dependencies. Such dynamic nonlinear coupling effects present a challenge for schedulers, which must construct cross-dimensional relationship models in dynamic, coupled environments to address local performance bottlenecks caused by resource contention.

\begin{figure}[!t]
\centering
\includegraphics[width=0.4\textwidth]{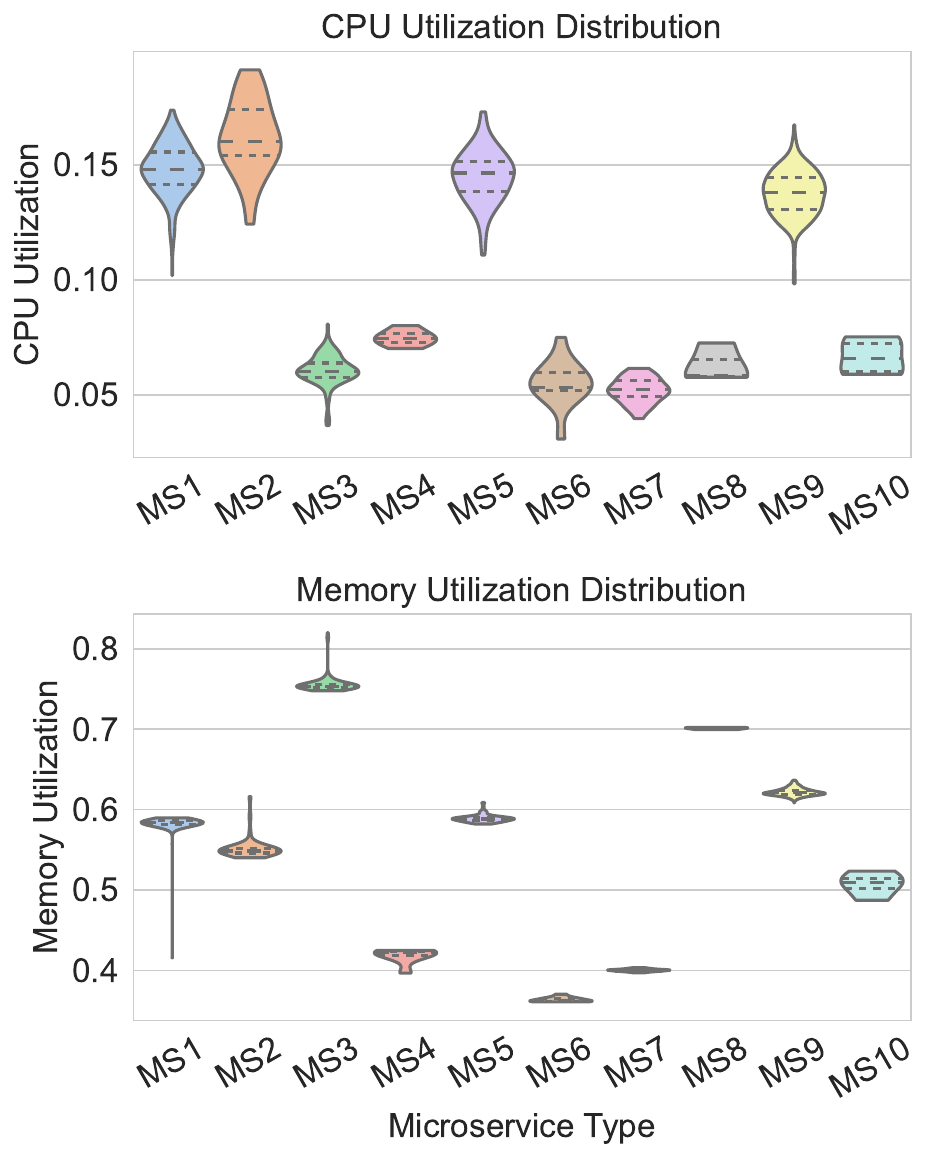}
\caption{Resource utilization distributions of microservice instances for 10 randomly selected microservice types, based on real-world production data from Alibaba.}
\label{fig2}
\end{figure}

{\bf Multi-dimensional heterogeneity of microservice resource requests.} As illustrated by Figs. \ref{fig1} and \ref{fig2}, there exists both spatial heterogeneity, where resource demands vary significantly among different microservices or instances and temporal dynamics, where the resource requirements of a single instance fluctuate in response to workload variations \cite{ref25} \cite{ref35}. When such heterogeneous instances are co-located on shared nodes, their diverse demands interact in complex ways, further amplifying heterogeneity at the node level. This multi-dimensional heterogeneity poses a challenge for schedulers, which must improve their generalization and adaptability to time-varying workloads in spatiotemporally heterogeneous environments, in order to achieve dynamic resource adaptation across diverse microservice types.

Traditional approaches, exemplified by heuristic algorithms, primarily rely on static modeling and empirically defined rules, making them poorly adaptable to dynamic environments and therefore often ineffective in dealing with the above challenges \cite{ref26} \cite{ref27}. Recent research has explored scheduling approaches based on deep reinforcement learning (DRL), which have demonstrated strong adaptability in dynamic cloud environments and are considered promising for future intelligent resource management systems \cite{ref14} \cite{ref30} \cite{ref31} \cite{ref33} \cite{ref36}. However, existing DRL-based approaches typically model the environment state directly using neural networks without explicitly capturing the complex characteristics of microservice systems. As a result, they often struggle to effectively address critical challenges including resource imbalance across nodes, nonlinear coupling among resource dimensions, and the heterogeneity of resource requests. These approaches lack the capability to adaptively extract structured and informative representations from raw system states, which limits their ability to generalize under dynamic and complex conditions. Motivated by these limitations, we propose to augment DRL with representation learning that learns compact and informative embeddings of the system state, enabling more effective scheduling policy learning under highly dynamic and heterogeneous microservice environments.

\noindent \textbf{Our work.} To effectively address the aforementioned key challenges and overcome the limitations of existing DRL-based scheduling approaches, we propose MCRL, a novel multi-resource cross-attention-based representation learning approach and MCRL2, a novel multi-resource cross-attention-based representation learning-augmented reinforcement learning approach for microservice scheduling.

To mitigate the fluctuating imbalance of multi-dimensional resources across nodes, MCRL introduces a cross-node multi-resource cross-attention mechanism that maps node-level resource vectors into a high-dimensional space and adaptively adjusts to non-stationary load fluctuations, thereby capturing dynamic correlations of resource features across nodes. By aggregating the resulting embeddings into compact cluster-level features encoding the global load distribution, the scheduler can maintain stable load balancing under dynamic workloads.

To capture the time-varying and nonlinear dependencies among resource dimensions, MCRL computes cross-resource attention weights for each node, explicitly modeling dynamic interdependencies between resources and capturing their interactions. Residual connections preserve the original resource embeddings to prevent information loss, enabling the scheduler to adaptively adjust its policy and alleviate local performance bottlenecks.

To handle the multi-dimensional heterogeneity of microservice resource requests, MCRL integrates node-level aggregated representations with the spatiotemporal heterogeneous features of microservice instances. Through feature fusion and nonlinear mapping, it jointly encodes the matching degree between microservice demands and resource capacities, as well as the affinity between service types and resource types, thereby enhancing the scheduler’s generalization capability across diverse and dynamic service scenarios.

To tackle the challenges of modeling high-dimensional heterogeneous states, mitigating value estimation bias, and stabilizing training under dynamic resource conditions, MCRL2 constructs a novel dual-stream soft actor-critic architecture that leverages MCRL representations. This design enhances the representation capacity of both policy and value networks, improves training stability through decoupled optimization and representation-guided exploration, and yields more reliable and efficient scheduling decisions. Notably, MCRL2 overcomes key limitations of existing DRL-based approaches in modeling complex resource interactions and state representation learning.

\noindent \textbf{Contributions.} The main contributions of this paper include:	
\begin{itemize}

\item{{\bf Representation Learning:} We propose MCRL, a novel multi-resource cross-attention-based representation learning approach that learns informative feature representations capturing the interactions across nodes, resources, and microservices (see Sec.~\ref{sec-MCRL}).}

\item{{\bf Reinforcement Learning:} We propose MCRL2, a novel reinforcement learning approach augmented with multi-resource cross-attention-based representation learning for microservice scheduling, which addresses the limitations of state representations and enables efficient and stable scheduling decisions (see Sec.~\ref{sec-MCRL2}).}

\item{{\bf Implementation \& Evaluation:} We conduct extensive experiments on the microservice dataset collected from the real production cluster, the results under different workload patterns show that the proposed scheduling approach outperforms than the baselines (see Sec.~\ref{sec-Eval}).}

\end{itemize}

The rest of the paper is organized as follows. Section 2 reviews related work. Section 3 defines the problem in detail. Section 4 describes the proposed multi-resource cross-attention-based representation learning approach. Section 5 introduces the reinforcement learning-based scheduling approach augmented with the representation learning. Section 6 presents the detail of performance evaluation. Finally, section 7 concludes the paper.

\section{RELATED WORK}

\noindent {\bf{Heuristic Approaches.}} Gu \emph {et al}. \cite{ref4} develop a layer-aware microservice placement and request scheduling algorithm based on an iterative greedy strategy with approximation guarantees, aiming to improve placement efficiency and service throughput. He \emph {et al}. \cite{ref21} propose a two-stage iterated greedy optimization algorithm for microservice placement in cloud-edge environments, which improves response time and execution efficiency over existing approaches. Ding \emph {et al}. \cite{ref23} propose an improved genetic algorithm for Kubernetes-oriented microservice placement with dynamic resource allocation, achieving higher throughput and lower cost compared to baselines. Kumar \emph {et al}. \cite{ref24} propose a QoS-aware resource scheduling model using a fine-tuned sunflower whale optimization algorithm for microservice applications, demonstrating improved performance over baselines in reducing delay, resource consumption, and service cost.

\noindent {\bf{DRL-based Approaches.}} Santos \emph {et al}. \cite{ref1} introduce a dynamic load balancer for microservice, which employs DRL including PPO, A2C, and DQN, along with DeepSets, to minimize latency and ensure load balancing in dynamic environments. Lv \emph {et al}. \cite{ref2} design a Deep Q Learning-based algorithm for addressing a multi-objective microservice deployment challenge in edge computing, aiming to reduce communication costs and enhance load balancing across nodes. Jian \emph {et al}. \cite{ref3} present a Deep Q Network-enhanced scheduler named DRS for microservice, which formulates scheduling as a Markov decision process to improve resource utilization and load balance. Wang \emph {et al}. \cite{ref5} present a microservice orchestration approach based on the delay-aware reward scaling proximal policy optimization reinforcement learning algorithm, which jointly optimizes deployment and routing to minimize delay and energy consumption. Boudieb \emph {et al}. \cite{ref6} develop a microservice instance selection approach based on double deep Q-Network to generate delay-aware service plans, effectively reducing deadline violation and improving load balancing. Wang \emph {et al}. \cite{ref10} propose a soft actor-critic-based scheduling approach that enhances load balancing in cloud computing services under dynamic workloads, demonstrating enhanced load-balancing effectiveness over baseline approaches. Table~\ref{tab:table_drl} summarizes and compares the DRL-based microservice scheduling approaches, highlighting their DRL algorithms, optimization objectives, and the state representation modeling strategies.

\begin{table*}[!t]
\caption{Comparison of Deep Reinforcement Learning-Based Scheduling Approaches for Microservice Systems.\label{tab:table_drl}}
\centering
\begin{tabular}{p{0.6in}<{\raggedright}p{1.0in}<{\raggedright}p{1.5in}<{\raggedright}p{3.0in}<{\raggedright}}
\hline
\textbf{References} & \textbf{DRL Algorithm} & \textbf{Optimization Objective} & \textbf{State Representation Modeling} \\
\hline
\\
Santos \emph {et al}. \cite{ref1} & PPO / A2C / DQN & Latency, load balancing & Generic feature vectors with standard NN encoders; no explicit modeling of cross-node/resource/service interactions \\
Lv \emph {et al}. \cite{ref2} & DQN & Communication cost, load balancing & Same as above \\
Jian \emph {et al}. \cite{ref3} & DQN & Average resource utilization, the degree of load balance & Same as above \\
Wang \emph {et al}. \cite{ref5} & PPO & Average delay, energy consumption & Same as above \\
Boudieb \emph {et al}. \cite{ref6} & Double DQN & Deadline violation, load balancing degree & Same as above \\
Wang \emph {et al}. \cite{ref10} & SAC & Variance of load fluctuation, task response time & Same as above \\
\bf{Our work} & \textbf{MCRL-augmented dual-stream soft actor-critic architecture} & \textbf{Load balancing, scheduling succcess rate} & \textbf{Multi-resource cross-attention representation learning; explicitly models cross-node correlations, cross-resource dependencies, and microservice-instance heterogeneity} \\
\hline
\end{tabular}
\end{table*}

\section{Problem Definition and Framework Overview}

\subsection{Problem Definition}

\noindent {\bf{System model.}} We define a cloud computing cluster as a finite set $\mathcal{N}=\{\mathcal{N}_1,\dots,\mathcal{N}_i,\dots,\mathcal{N}_{|\mathcal{N}|}\}$, where $\mathcal{|N|}$ represents the number of nodes. $\mathcal{RC}_i=[\mathcal{RC}_i^1,\dots,\mathcal{RC}_i^j,\dots,\mathcal{RC}_i^J ]$ and $\mathcal{RU}_i=[\mathcal{RU}_i^1,\dots,\mathcal{RU}_i^j,\dots,\mathcal{RU}_i^J ]$ respectively represents the resource capacity vector and resource utilization vector of $i$-th node, where $\mathcal{RC}_i^j$ represents the $j$-th resource dimension of $\mathcal{N}_i$, $J$ represents the resource dimension number of $\mathcal{N}_i$. The microservice applications running on the cluster is considered as a finite set $\mathcal{MS}=\{\mathcal{MS}_1,\dots,\mathcal{MS}_k,\dots,\mathcal{MS}_\mathcal{|MS|}\}$, where $\mathcal{|MS|}$ represents the number of microservices. The microservices are all containerized and each microservice instance of a microservice runs in an isolated container. $\mathcal{MI}_k=\{\mathcal{MI}_{k1},\dots,\mathcal{MI}_{kl},\dots,\mathcal{MI}_{k{|\mathcal{MI}_k|}}\}$ represents the instance set of $k$-th microservice, where $\mathcal{MI}_{kl}$ represents the $l$-th instance of $\mathcal{MS}_k$, $|\mathcal{MI}_k|$ represents the instance number of $\mathcal{MS}_k$. For $\mathcal{MI}_{kl}$, we define $\mathcal{RR}_{kl}=[\mathcal{RR}_{kl}^1,\dots,\mathcal{RR}_{kl}^j,\dots,\mathcal{RR}_{kl}^J]$ as the resource request vector of $\mathcal{MI}_{kl}$, where $\mathcal{RR}_{kl}^j$ represents the $j$-th resource request dimension of $\mathcal{MI}_{kl}$.

Moreover, a function $\mathcal{{F}_{RC}}$ mapping each node to its resource capacity set is defined as Eq. \eqref{eq1}.

\begin{equation}
\label{eq1}
  \mathcal{{F}_{RC}} : \mathcal{N} \rightarrow \mathbb{R}_+^J, \quad \mathcal{{F}_{RC}}(\mathcal{N}_i) = \mathcal{RC}_i
\end{equation}

A function $\mathcal{{F}_{RU}}$ mapping each node to its resource utilization set is defined as Eq. \eqref{eq2}.

\begin{equation}
\label{eq2}
  \mathcal{{F}_{RU}} : \mathcal{N} \rightarrow \mathbb{R}_+^J, \quad \mathcal{{F}_{RU}}(\mathcal{N}_i) = \mathcal{RU}_i
\end{equation}

Similarly, a function $\mathcal{{F}_{RR}}$ mapping each microservice instance to its resource request set is defined as Eq. \eqref{eq3}

\begin{equation}
\label{eq3}
  \mathcal{F}_{RR} : \bigcup_{k=1}^{\lvert \mathcal{MS} \rvert} \mathcal{MI}_k \rightarrow \mathbb{R}_+^J, \quad \mathcal{F}_{RR}(\mathcal{MI}_{kl}) = \mathcal{RR}_{kl}
\end{equation}

Therefore, based on the above definition, the system model $\mathcal{SM}$ is optimally structured as a heterogeneous, ordered, and extensible tuple, which is formulated as follows.

\begin{equation}
\label{eq4}
 \mathcal{SM} = \left( 
{\mathcal{N}},\ 
{\mathcal{F}}_{\mathcal{RC}},\ 
{\mathcal{F}}_{{\mathcal{RU}}},\ 
{\mathcal{MS}},\ 
\left\{ {\mathcal{MI}}_k \right\}_{k=1}^{\lvert {\mathcal{MS}} \rvert},\ 
{\mathcal{F}}_{\mathcal{RR}},\ 
J 
\right)
\end{equation}

\noindent where $J \in \mathbb{Z}^+$.

\noindent {\bf{Scheduling model.}} The service applications are deployed on cloud platforms to provide continuous and reliable online services to users. As user requests fluctuate, the autoscaling mechanism dynamically adjusts the number of microservice instances by creating and terminating instances in response to workload variations. Newly created microservice instances must be scheduled onto appropriate nodes to ensure efficient resource utilization and service performance.

Microservice scheduling is essentially a multi-objective, multi-dimensional resource-constrained combinatorial optimization problem. It requires identifying the optimal (or near-optimal) allocation from all possible scheduling schemes. However, the solution space grows exponentially with the number of microservice instances and nodes, making it an NP-hard problem. We define a binary decision variable $x_{kl}^i \in \{0, 1\}$ as follows.

\begin{equation}
\label{eq5}
  x_{kl}^i =
\begin{cases}
1, & \text{if } \mathcal{MI}_{kl} \text{ is assigned to } \mathcal{N}_i \\
0, & \text{otherwise}
\end{cases} 
\end{equation}

The scheduling process is constrained by multi-dimensional resources capacity. For each node $\mathcal{N}_i$ and each resource dimension j, the sum of the resource demands of all instances assigned to node $\mathcal{N}_i$ have to meet the constraint as Eq. \eqref{eq6}. 

\begin{equation}
\label{eq6}
\begin{gathered}
\sum_{k=1}^{\lvert \mathcal{MS} \rvert} 
\sum_{l=1}^{\lvert \mathcal{MI}_k \rvert} 
x_{kl}^i \cdot \mathcal{RR}_{kl}^j 
\leq \mathcal{RC}_i^j, 
\\
\forall i \in \{1, \dots, \lvert \mathcal{N} \rvert\},\quad 
\forall j \in \{1, \dots, J\}
\end{gathered}
\end{equation}

Besides, each microservice instance must and can only be assigned to exactly one node, the constraint is defined as Eq. \eqref{eq7}. 

\begin{equation}
\label{eq7}
\begin{gathered}
\sum_{i=1}^{\lvert \mathcal{N} \rvert} x_{kl}^i = 1,
\\
\forall k \in \{1, \dots, \lvert \mathcal{MS} \rvert\},\quad 
\forall l \in \{1, \dots, \lvert \mathcal{MI}_k \rvert\}
\end{gathered}
\end{equation}

\begin{figure*}[!t]
\centering
\includegraphics[width=0.9\textwidth]{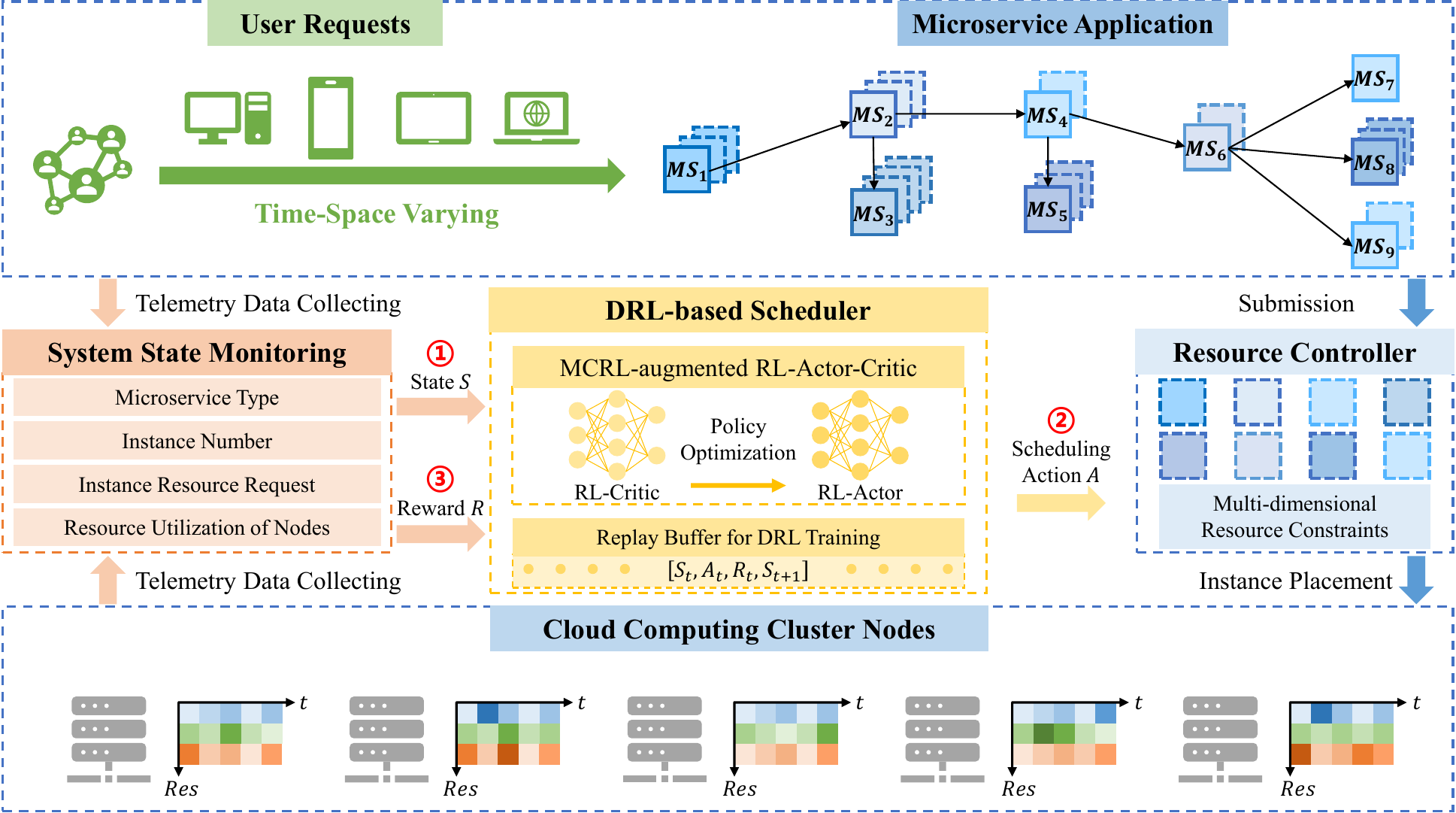}
\caption{Framework Overview of MCRL2 for Microservice Scheduling in the Cloud.}
\label{fig_framework}
\end{figure*}

\noindent {\bf{Multi-objective optimization problem.}} We take load balance of cluster and scheduling success rate of microservice instance into account. 

$\mathcal{ARU}
=[\mathcal{ARU}_1,\dots,{ARU}_j,\dots,{ARU}_J]$ is defined as the average resource utilization vector of the cluster, where ${ARU}_j$ represents the $j$-th average resource utilization. ${ARU}_j$ is calculated by the Eq. \eqref{eq8}.

\begin{equation}
\label{eq8}
   \mathcal{ARU} = \frac{\sum_{i=1}^{\lvert \mathcal{N} \rvert} \mathcal{RU}_i^j}{\lvert \mathcal{N} \rvert}
\end{equation}

Then, the load balancing degree of the $j$-th resource dimension of the cluster $\mathcal{LB}_j$ is calculated by the Eq. \eqref{eq9}.

\begin{equation}
\label{eq9}
   \mathcal{LB}_j = \sqrt{\frac{\sum_{i=1}^{\lvert \mathcal{N} \rvert} \left( \mathcal{RU}_i^j - \mathcal{ARU}_j \right)^2}{\lvert \mathcal{N} \rvert}}
\end{equation}

Therefore, the first optimization objective considering load balancing can be defined as follows.

\begin{equation}
\label{eq10}
   \mathcal{LB} = \sum_{j=1}^{J} \omega_j \cdot \mathcal{LB}_j
\end{equation}

\noindent where $\omega_j$ the weight of the degree of load balancing for the $j$-th dimension of the resource.

Next, we introduce the optimization objective of microservice scheduling success rate as in Eq. \eqref{eq11}.

\begin{equation}
\mathcal{MSSR} = 
\frac{
\begin{aligned}
& \sum_{k=1}^{\lvert \mathcal{MS} \rvert} \sum_{l=1}^{\lvert \mathcal{MI}_k \rvert}
\mathbb{I} \left( \sum_{i=1}^{\lvert \mathcal{N} \rvert} x_{kl}^i = 1 \right) \\
& \quad \cdot \mathbb{I} \left( \forall j,\ \sum_{k',l'} x_{k'l'}^i \cdot \mathcal{RR}_{k'l'}^j \leq \mathcal{RC}_i^j \right)
\end{aligned}
}{
\sum_{k=1}^{\lvert \mathcal{MS} \rvert} \lvert \mathcal{MI}_k \rvert
}
\label{eq11}
\end{equation}

\noindent where $\mathbb{I}$ represents the indicator function, which returns 1 if a given condition is true and 0 otherwise. 

Overall, the multi-objective optimization problem addressed in this study can be formally formulated as follows.

\begin{equation}
\label{eq12}
\begin{array}{c}
\min \left( \mathcal{LB},\ 1 - \mathcal{MSSR} \right)
\\[0.5em]
\text{s.t.} \quad \text{Eq. \eqref{eq6}},\quad \text{Eq. \eqref{eq7}},
\quad x_{kl}^i \in \{0, 1\}
\end{array}
\end{equation}

\subsection{Framework Overview}

The framework overview of the proposed MCRL2 for microservice scheduling in the cloud is illustrated in Fig. \ref{fig_framework}. To cope with dynamically changing user requests, the cloud platform adjusts the number of microservice instances through autoscaling. Each instance has distinct resource demands and is placed on appropriate cluster nodes via the Resource Controller. MCRL2 implements the scheduler following the DRL paradigm. Specifically, the System State Monitoring module continuously collects telemetry data from both the application and resource layers to construct the environment state. Based on the state, the scheduler makes decisions to place submitted instances on appropriate nodes, thereby optimizing resource utilization and load balancing across the cluster. The reward signal is derived from the updated system state, measuring the immediate return of each scheduling decision with respect to the system optimization objectives. This interaction is formalized as a Markov decision process, where state–action–reward–next state tuples are sequentially formed into experience trajectories to train the DRL agent. In particular, MCRL2 leverages the proposed RL-Actor-Critic architecture to optimize scheduling policy learning through representation learning-augmented reinforcement learning.

\section{Multi-resource Cross-Attention Mechanism-based Representation Learning Approach}
\label{sec-MCRL}

\subsection{Multi-resource Cross-Attention Mechanism}

We define the resource utilization matrix $\mathcal{RU}$ of $\mathcal{N}$ as Eq. \eqref{eq13}. The input of MCAM is defined as the resource-specific embedding $E \in \mathbb{R}^{\lvert \mathcal{N} \rvert \times J \times d}$, where $E^j$ is the embedding of the resource corresponding to the dimension of the $j$-th resource of the cluster.

\begin{equation}
\label{eq13}
   \mathcal{RU} = [\mathcal{RU}_1, \dots , \mathcal{RU}_i, \dots, \mathcal{RU}_{\lvert \mathcal{N} \rvert}]^T
\end{equation}

\noindent {\bf{Multi-resource Linear Projection.}} For each resource dimension $j \in \{1, \dots, J\}$, MCAM performs an independent linear transformation, which generates a triplet of Query, Key and Value \cite{ref32}.

\begin{equation}
\label{eq14}
\begin{aligned}
Q^j &= W_Q^j \cdot E^j + b_Q^j, \\
K^j &= W_K^j \cdot E^j + b_K^j, \\
V^j &= W_V^j \cdot E^j + b_V^j
\end{aligned}
\end{equation}

\noindent where $W_Q^j,\ W_K^j,\ W_V^j \in \mathbb{R}^{d \times d}$ is the trainable projection matrices corresponding to resource $j$. $d$ represents the embedding space dimensionality. $b_Q^j,\ b_K^j,\ b_V^j \in \mathbb{R}^d$ are bias terms.

\noindent {\bf{Cross-Resource Attention Weights.}} The cross-resource attention tensor $A \in \mathbb{R}^{\lvert \mathcal{N} \rvert \times J \times J}$ encodes dynamic directional resource coordination policies, where each element $A_i^{j, p}$ encodes a learnable decision rule. This rule determines the extent to which the state of the $j$-th resource type influences the allocation policy of the $p$-th resource type for the node $\mathcal{N}_i$, based on real-time telemetry data and global scheduling objectives. $A_i^{j, p}$ is calculated as follows.

\begin{equation}
\label{eq15}
\begin{gathered}
A_i^{(j,p)} = \mathrm{Softmax} \left( \frac{Q_i^j \cdot (K_i^p)^\mathrm{T}}{\sqrt{d}} \right),
\\
\forall i \in \{1, \dots, \lvert \mathcal{N} \rvert\},\quad 
\forall j, p \in \{1, \dots, J\}
\end{gathered}
\end{equation}

\noindent {\bf{Dynamic Resource Fusion.}} MCAM fuses the resource value vectors of each resource dimension based on the attention weights.

\begin{equation}
\label{eq16}
\begin{gathered}
Z_i^j = \sum_{p=1}^{J} A_i^{(j,p)} \cdot V_i^p,
\\
\forall i \in \{1, \dots, \lvert \mathcal{N} \rvert\},\quad 
\forall j \in \{1, \dots, J\}
\end{gathered}
\end{equation}

\noindent where $Z \in \mathbb{R}^{\lvert \mathcal{N} \rvert \times J \times d}$ represents the fused feature representations across resource dimensions.

\noindent {\bf{Residual Connection.}} MCAM employs residual connections to preserve the original input information and enhance training stability \cite{ref34}. 

\begin{equation}
\label{eq17}
\begin{gathered}
Y^j = \text{LayerNorm}(Z^j + E^j),
\\
\forall j \in \{1, \dots, J\}
\end{gathered}
\end{equation}

\noindent where $Y^j \in \mathbb{R}^{\lvert \mathcal{N} \rvert \times d}$ encodes the dynamically enhanced representations of each resource dimension for the cluster, which retains the original resource utilization information while incorporating cross-resource dependencies. Moreover, $Y^j$ provides a fine-grained resource-aware representations for subsequent node-level feature fusion.

\subsection{Multi-resource Cross-Attention Mechanism-based Representation Learning}

\begin{figure}[!t]
\centering
\includegraphics[width=0.45\textwidth]{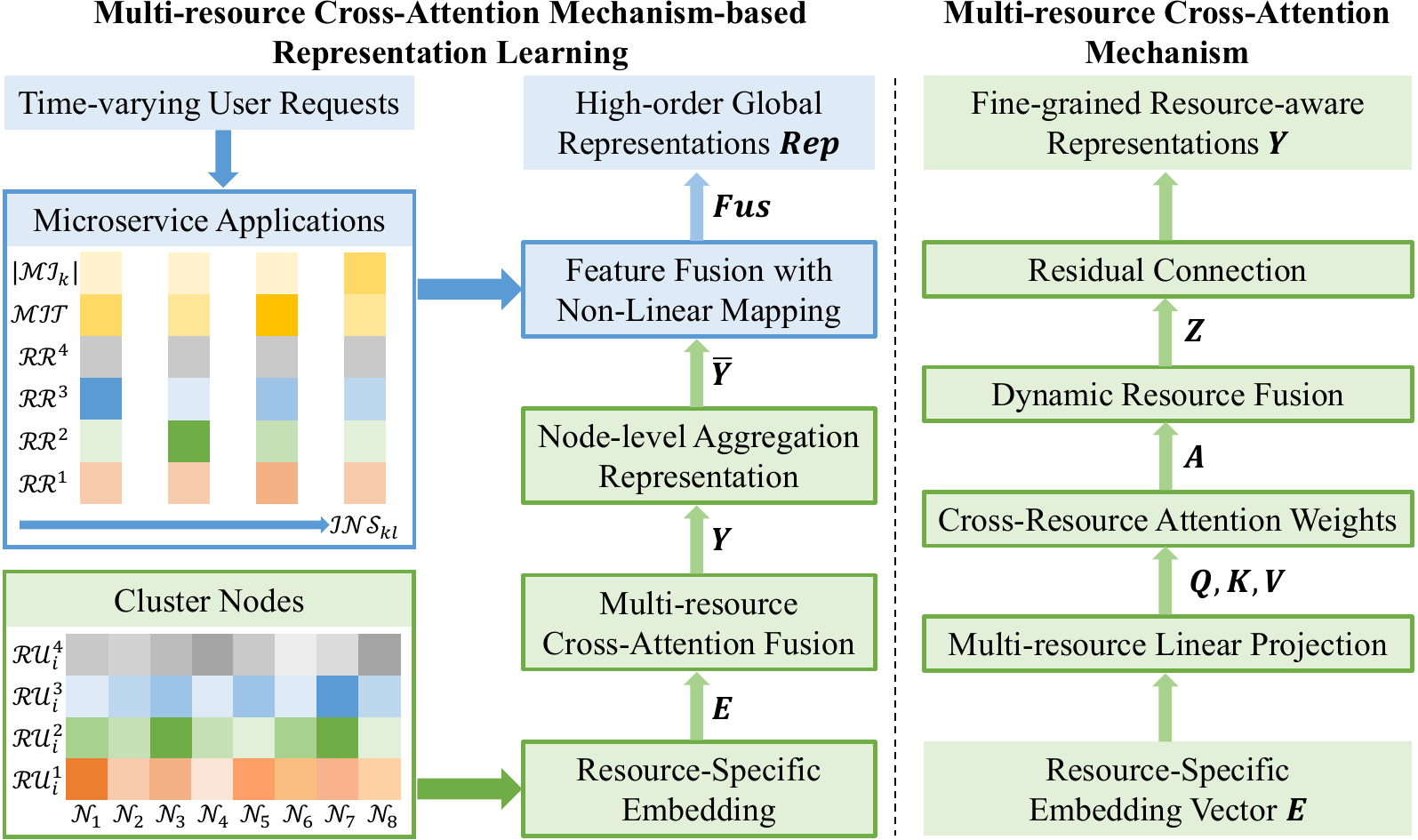}
\caption{Workflow of Multi-resource Cross-Attention Mechanism-based Representation Learning and Multi-resource Cross-Attention Mechanism.}
\label{mcrl_mcam}
\end{figure}

Based on multi-resource cross-Attention mechanism, beyond considering the resource information of nodes, we further incorporate the resource requests and types of microservice instances. We propose a novel multi-resource cross-attention mechanism-based representation learning approach, namely MCRL. The workflow of MCRL and MCAM is shown in Fig. \ref{mcrl_mcam}. We first define the information vector of microservice instance $\mathcal{MI}_{kl}$ as follows.

\begin{equation}
\label{eq18}
\mathcal{INS}_{kl} = [\mathcal{MIT}_{kl},\ \lvert \mathcal{MI}_k \rvert,\ \mathcal{RR}_{kl}^1,\ \dots,\ \mathcal{RR}_{kl}^j,\ \dots,\ \mathcal{RR}_{kl}^J]
\end{equation}

\noindent where $\mathcal{MIT}_{kl}$ indicates the running time of $\mathcal{MI}_{kl}$.

\noindent {\bf{Resource-Specific Embedding.}} MCRL first decouples resource vectors of $\mathcal{RU}$ across different dimensions and maps them into high-dimensional embedding spaces separately.

\begin{equation}
\label{eq19}
\begin{gathered}
E^j = W_E^j \cdot \mathcal{RU}_{:}^j + b^j,
\forall j \in \{1, \dots, J\}
\end{gathered}
\end{equation}

\noindent where $W_E^j \in \mathbb{R}^{1 \times d}$ is the resource-specific embedding matrix, $\mathcal{RU}_{:}^j$ represents the resource utilization vector of the $j$-th dimension for the cluster.

\noindent {\bf{Multi-resource Cross-Attention Fusion.}} For the $E^j,\, \forall j \in \{1, \dots, J\}$, MCRL stacks resource embeddings from different dimensions and processes them using multi-resource cross-attention mechanism.

\begin{equation}
\label{eq20}
\begin{gathered}
Y = \text{MCAM} \left( \text{Stack}(E^1, \dots, E^j, \dots, E^J) \right)
\end{gathered}
\end{equation}

\noindent where $Y \in \mathbb{R}^{\lvert \mathcal{N} \rvert \times J \times d}$ is the tensor obtained by stacking the $Y^j$ computed by MCAM. $Y$ represents the global resource representations of the fused dynamic correlation information across resource dimensions. By utilizing MCAM, MCRL not only acquires fine-grained resource interaction features but also explicitly encodes dynamic coordination policies across multiple resource dimensions in $Y$.

\noindent {\bf{Node-level Aggregation Representation.}} MCRL applies average pooling to the fused dynamic correlation information $Y$ across resource dimensions at the node level to compress feature representations.

\begin{equation}
\label{eq21}
\begin{gathered}
\overline{Y}_i = \frac{1}{J} \sum_{j=1}^{J} Y_i^j,
\forall i \in \{1, \dots, \lvert \mathcal{N} \rvert\}
\end{gathered}
\end{equation}

\noindent where $Y_i^j$ denotes the dynamic correlation feature of $i$-th node in $j$-th resource dimension. $\overline{Y}_i$ is the multi-resource aggregated representation of node $\mathcal{N}_i$, obtained by averaging the features across all resource dimensions, integrating dynamic interaction information across resource dimensions. It encodes the overall workload state of node $\mathcal{N}_i$, considering the collaborative or competitive relationships between different resources.

\noindent {\bf{Feature Fusion with Non-Linear Mapping.}} The feature representations across resource dimensions at the node level are flattened and concatenated with microservice instance information.

\begin{equation}
\label{eq22}
\begin{gathered}
{Fus} = \text{concat} \left( \text{flat} \left( \left[ \overline{Y}_1, \dots, \overline{Y}_i, \dots, \overline{Y}_{\lvert \mathcal{N} \rvert} \right] \right),\ \mathcal{INS}_{kl} \right)
\end{gathered}
\end{equation}

where $Fus$ represents the joint feature of globally aggregated node-level features and microservice instances within the cluster, providing comprehensive contextual information for microservice scheduling decisions.
Finally, $Fus$ is encoded into the final feature representations for microservice scheduling through hidden layers.

\begin{equation}
\label{eq23}
\begin{gathered}
{Rep} = \text{ReLU} \left( W_{fc} \cdot Fus + b_{fc} \right)
\end{gathered}
\end{equation}

where $Rep$ encodes the high-order global representations learned by MCRL for microservice scheduling decisions, encompassing information about the cluster nodes, microservices, and microservice instances. The entire workflow of MCRL is illustrated in Algorithm \ref{alg:alg1}. 

\begin{algorithm}[H]
\caption{Pseudocode of MCRL}\label{alg:alg1}
\begin{algorithmic}[1]
\STATE {\textbf{Input:}} Resource utilization matrix $\mathcal{RU}$, information vector $\mathcal{INS}_{kl}$ of microservice instance $\mathcal{MI}_{kl}$.
\STATE {\textbf{Output:}} Learned high-order global representations $Rep$.
\STATE Initialize multi-resource cross-attention mechanism model \text{MCAM} with parameters $\theta_{MCAM}$.
\STATE {\textbf{for}} $t=1$ to $T$ {\textbf{do}}
\STATE \hspace{0.5cm} Compute resource-specific embedding $E$ by Eq. \eqref{eq19} \\
        \hspace{0.5cm} based on $\mathcal{RU}$.
\STATE \hspace{0.5cm} Compute global resource representations $Y$ by Eq. \eqref{eq20} \\
        \hspace{0.5cm} based on \text{MCAM} and $E$.
\STATE \hspace{0.5cm} Compute multi-resource aggregated representations $\overline{Y}$ \\
        \hspace{0.5cm} of all cluster nodes by Eq. \eqref{eq21} based on $Y$.
\STATE \hspace{0.5cm} Obtain joint feature $Fus$ of nodes and microservice \\
         \hspace{0.5cm} instances by Eq. \eqref{eq22} based on $\overline{Y}$ and $\mathcal{INS}_{kl}$.
\STATE \hspace{0.5cm} Obtain $Rep$ by Eq. \eqref{eq23} based on $Fus$.
\STATE \textbf{end for}
\end{algorithmic}
\label{alg1}
\end{algorithm}

\section{Multi-resource Cross-attention-based Representation Learning-augmented Reinforcement Learning for Microservice Scheduling}
\label{sec-MCRL2}

\subsection{Representation Learning-augmented Reinforcement Learning Model for Microservice Scheduling}

The microservice scheduling in cloud can be formulated as a Markov Decision Process (MDP) defined by $\langle \mathcal{S}, \mathcal{A}, \mathcal{P}, \mathcal{R}, \gamma \rangle$.

\begin{table}[!t]
\caption{The node-specific, microservice-specific and instance -specific telemetry data. \label{tab:table1}}
\centering
\begin{tabular}{c c}
\hline
Level & Telemetry Data \\
\hline
Cluster Node & \makecell{Resource utilization: \\ $[\mathcal{RU}_1, \dots, \mathcal{RU}_i, \dots, \mathcal{RU}_{\lvert \mathcal{N} \rvert}]$} \\
Microservice & \makecell{Microservice type and instance number: \\ $[\mathcal{MIT}_{kl},\ \lvert \mathcal{MI}_k \rvert]$} \\
Microservice Instance & \makecell{Instance resource request: \\ $[\mathcal{RR}_{kl}^1,\,\dots,\,\mathcal{RR}_{kl}^j,\,\dots,\,\mathcal{RR}_{kl}^J]$} \\
\hline
\end{tabular}
\end{table}

{\bf{State Space $\mathcal{S}$ for Microservice Scheduling.}} $\mathcal{S}$ is defined as the set of all possible environmental states. Scheduling agent collects the node-specific, microservice-specific and instance-specific telemetry data shown as Table \ref{tab:table1} and learns policy for microservice scheduling. The environmental state observed by the RL agent at timestep $t$ is defined as $s_t \in \mathcal{S}$. The state size is $|\mathcal{N}| + 2 + J$.

\begin{equation}
\label{eq24}
\begin{split}
s_t = \left[ \mathcal{RU}_1, \dots, \mathcal{RU}_i, \dots, \mathcal{RU}_{\lvert \mathcal{N} \rvert},\ 
\mathcal{MIT}_{kl},\ \lvert \mathcal{MI}_k \rvert, \right. \\
\left. \mathcal{RR}_{kl}^1, \dots, \mathcal{RR}_{kl}^j, \dots, \mathcal{RR}_{kl}^J \right]
\end{split}
\end{equation}

{\bf{Action Space $\mathcal{A}$ for Microservice Scheduling.}} $\mathcal{A}$ is defined as the set of all possible actions can be executed in the environment. $a_t \in \mathcal{A}$ is defined as the action determined by the agent at timestep $t$.

\begin{equation}
\label{eq25}
A = \{1, \dots, \lvert \mathcal{N} \rvert\}
\end{equation}

{\bf{State Transition Probability $\mathcal{P}$ for Microservice Scheduling.}} $\mathcal{P}$ is defined as the probability of the state transition from $s$ to $s'$ after the action $a$ is executed.

\begin{equation}
\label{eq26}
\begin{gathered}
P(s' \mid s, a) = \Pr(s_{t+1} = s' \mid s_t = s,\, a_t = a),
\\
\text{s.t. } \sum_{s' \in \mathcal{S}} P(s' \mid s, a) = 1
\end{gathered}
\end{equation}

\noindent where $s,s' \in \mathcal{S}$ represent the current and next state respectively.

{\bf{Reward Function $\mathcal{R}$ for Microservice Scheduling.}} $\mathcal{R}$ is defined as the immediate reward received by the agent after the state transitioning from $s$ to $s'$ by executing action $a$. First, we define the reward function $\mathcal{R}_1$ when the microservice instance $\mathcal{MI}_{kl}$ is scheduled to the node $\mathcal{N}_i$ successfully.

\begin{equation}
\label{eq27}
\mathcal{R}_1 = 
\begin{cases}
2.0, & 
\begin{aligned}
&\text{if } \mathcal{ARU}_j - \mathcal{RU}_i^j \geq 0, \\
&\forall j \in \{1, \dots, J\}
\end{aligned}
\\
\sum_{j=1}^{J} \left( \mathcal{ARU}_j - \mathcal{RU}_i^j \right), & \text{otherwise}
\end{cases}
\end{equation}

When $\mathcal{MI}_{kl}$ failed to be scheduled to $\mathcal{N}_i$, we define reward function $\mathcal{R}_2=-1.0$. Combining $\mathcal{R}_1$ and $\mathcal{R}_2$, we define the comprehensive reward function $\mathcal{R}$ as Eq. \eqref{eq28}.

\begin{equation}
\label{eq28}
\mathcal{R} = 
\begin{cases}
-1.0, & \text{if } x_{kl}^i = 0 \\
\mathcal{R}_1, & \text{otherwise}
\end{cases}
\end{equation}

The reward function penalizes scheduling failures, thereby guiding the agent to learn how to allocate resources efficiently under capacity constraints. Additionally, by dynamically adjusting the reward based on the deviation between the node workload and the cluster average workload, the function further encourages the agent to select nodes that promote overall load balancing. This design effectively drives the agent towards developing an optimized strategy that addresses multiple objectives simultaneously.

{\bf{Discount Factor $\mathcal{\gamma}$ for Microservice Scheduling.}} $\gamma$ is used to balance the importance of immediate reward and future reward. The cumulative return $G_t$ is defined as follows. Maximizing $G_t$ is the core objective of agent, directly determining the optimization direction of the scheduling policy.

\begin{equation}
\label{eq29}
G_t = \sum_{k=0}^{\infty} \gamma^k R_{t+k+1}
\end{equation}

\subsection{Implementation Details of Multi-resource Cross-Attention-based Representation Learning-augmented Reinforcement Learning}

To address the limitations of existing reinforcement learning-based approaches in representing complex, high-dimensional system states and mitigating value estimation bias, we propose MCRL2. It integrates representation learning with the actor-critic framework and introduces a novel dual-stream architecture network RL-Actor-Critic.

{\bf{Dual-Stream Network Architecture.}} To effectively integrate representation learning with reinforcement learning based on the actor-critic architecture, we design the RL-Actor-Critic, consisting of RL-Actor and RL-Critic. RL-Actor is sequentially composed of RLA, a representation learning module based on MCRL, and a policy network actor. These components are responsible for extracting state representations and generating action decisions, respectively. Similarly, RL-Critic includes RLC, a representation learning module based on MCRL, and a Q-value network critic. RLC extracts key representations of state-action pairs, while the critic network evaluates action values and provides signals for policy optimization. This design enhances the representation capacity of both policy generation and value estimation, thereby improving the overall performance of reinforcement learning. 

{\bf{Representation-Guided Structured Exploration.}} MCRL2 incorporates an entropy regularization mechanism to constrain the exploration direction. More critically, MCRL2 leverages guidance from multi-resource cross-attention representation learning to prioritize exploration of critical nodes.

{\bf{Decoupled Optimization Pathways.}} The policy network and value network are updated independently, which effectively prevents training instabilities caused by parameter coupling.

We define the MCRL-based state encoder function RLA and RLC as follows. 

\begin{equation}
\label{eq30}
\begin{gathered}
{Rep}_t^{rla} = \text{RLA}(s_t;\ \theta^{rla}) \\
{Rep}_t^{rlc} = \text{RLC}(s_t;\ \theta^{rlc})
\end{gathered}
\end{equation}

\noindent where $s_t$ is defined in Eq. \eqref{eq24}, ${Rep}_t^{rla}$ and ${Rep}_t^{rlc}$ indicate the high-order representations for Actor and Critic learned by MCRL at timestep $t$, respectively. $\theta^{rla}$ and $\theta^{rlc}$ represent the independent parameters of RLA and RLC, respectively.

RL-Actor-Critic architecture combines MCRL with the soft actor-critic under the maximum entropy framework to learn the policy for microservice scheduling. The objective function is defined as follows.

\begin{equation}
\label{eq31}
\begin{aligned}
\mathcal{J}(\pi) =\ & \mathbb{E}_{(s_t,a_t) \sim \rho_\pi} \bigg[ \sum_{t=0}^{\infty} \gamma^t \big( r({Rep}_t^{rla}, a_t) \\
& + \alpha \mathcal{H}(\pi(\cdot \mid {Rep}_t^{rla})) \big) \bigg]
\end{aligned}
\end{equation}

\noindent where $\mathcal{H}(\pi(\cdot \mid {Rep}_t^{rla})) = -\mathbb{E}_{a \sim \pi} \left[ \log \pi(a \mid {Rep}_t^{rla}) \right]$ indicates the entropy of the policy in feature representation space. The randomness of the policy is based on the feature representations ${Rep}_t^{rla}$, making exploration more efficient. $\alpha$ is the temperature parameter that balances the trade-off between reward maximization and entropy regularization in the policy. And the optimal policy $\pi^*$ is defined as Eq. \eqref{eq32}.

\begin{equation}
\label{eq32}
\pi^* = \arg \max_{\pi} \mathcal{J}(\pi)
\end{equation}

\begin{figure}[!t]
\centering
\includegraphics[width=0.45\textwidth]{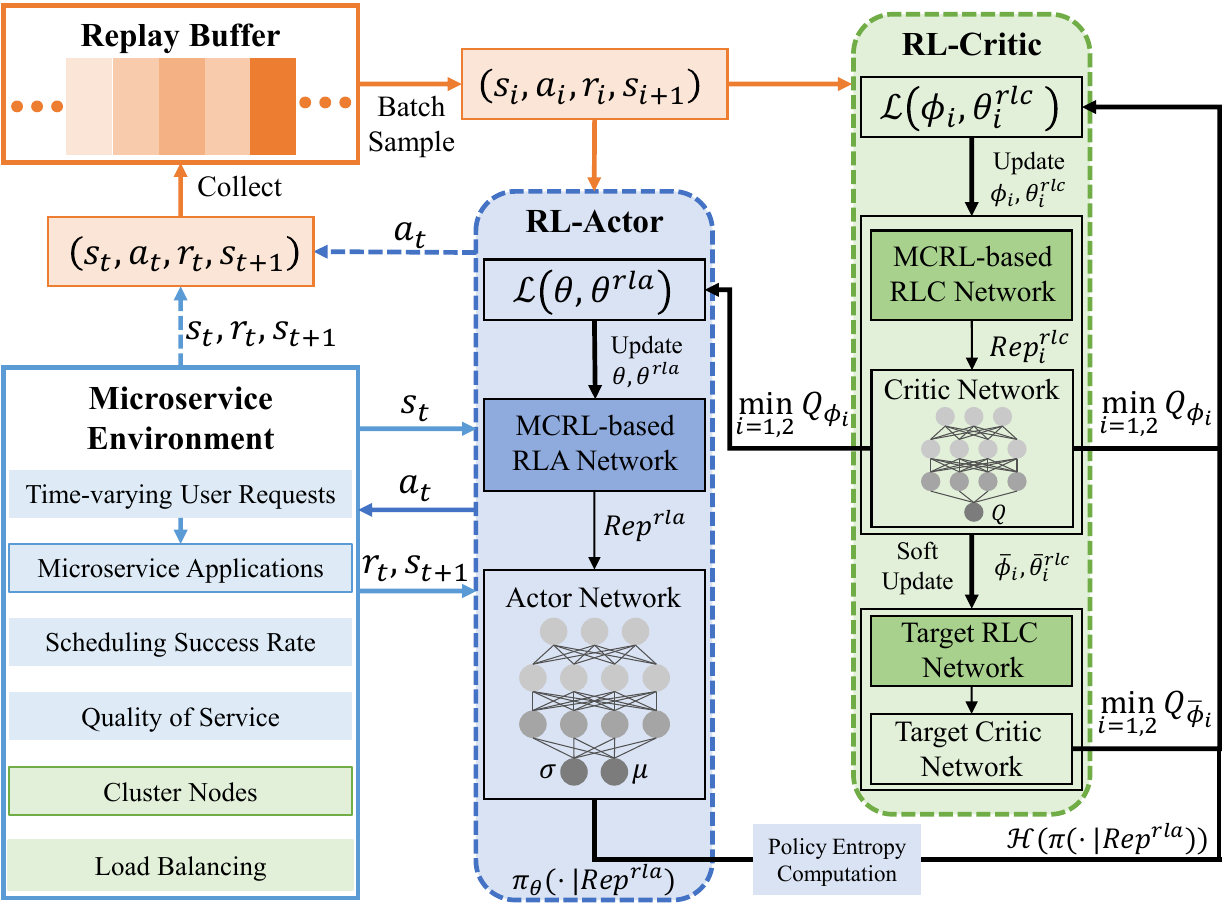}
\caption{Architecture of Multi-resource Cross-Attention-based Representation Learning-augmented Reinforcement Learning.}
\label{mcrl2}
\end{figure}

\begin{algorithm}[H]
\caption{Pseudocode of MCRL2}\label{alg:alg2}
\begin{algorithmic}[1]
\STATE {\textbf{Input:}} $\text{RLA}(\cdot; \theta^{rla})$ , $\pi_{\theta}$, $\text{RLC}(\cdot; \theta^{rlc}_i)$, $Q_{\phi_i}$, $i=1,2$.
\STATE {\textbf{Output:}} Learned microservice scheduling policy $\pi_\theta$.
\STATE Initialize temperature $\alpha$
\STATE Initialize interpolation $\tau$, replay buffer $\mathcal{B} \leftarrow \emptyset$
\STATE Initialize target network $\bar{\theta}^{rlc}_i \leftarrow \theta^{rlc}_i$, $\bar{\phi}_i \leftarrow {\phi}_i$, $i=1,2$
\STATE Initialize step size $\eta_\pi^1$, $\eta_\pi^2$, $\eta_Q^1$, $\eta_Q^2$, $\eta_\alpha$, $\tau^1$, $\tau^2$.
\STATE {\textbf{for}} $e=1$ to $MaxEpisode$ {\textbf{do}}
\STATE \hspace{0.5cm} {\textbf{for}} $t=1$ to $MaxEnvStep$ {\textbf{do}}
\STATE \hspace{1.0cm} Obtain $s_t$ from cloud microservice environment.
\STATE \hspace{1.0cm} Compute ${Rep}^{rla}_t$ by Algo. \ref{alg:alg1}: \\
        \hspace{1.0cm} ${Rep}_t^{rla} = \text{RLA}(s_t; \theta^{rla})$
\STATE \hspace{1.0cm} Sample action $a_t \sim \pi_\theta (\cdot \mid {Rep}^{rla}_t)$.
\STATE \hspace{1.0cm} Obtain next state $s_{t+1}$ after executing $a_t$.
\STATE \hspace{1.0cm} Calculate reward $r_t$ by Eq. \eqref{eq28}.
\STATE \hspace{1.0cm} Store experiences in replay buffer: \\
        \hspace{1.0cm} $\mathcal{B} \leftarrow \mathcal{B} \cup \{s_t,a_t,r_t,s_{t+1}\}$
\STATE \hspace{0.5cm} {\textbf{end for}}
\STATE \hspace{0.5cm} {\textbf{for}} each training step {\textbf{do}}
\STATE \hspace{1.0cm} Sample a batch of experiences from $\mathcal{B}$

\STATE \hspace{1.0cm} $\theta^{rla} \leftarrow \theta^{rla}-\eta_\pi^1 \nabla_{\theta^{rla}} \mathcal{L}(\theta,\theta^{rla})$
\STATE \hspace{1.0cm} $\theta \leftarrow \theta-\eta_\pi^2 \nabla_\theta \mathcal{L}(\theta,\theta^{rla})$
\STATE \hspace{1.0cm} $\theta^{rlc}_i \leftarrow \theta^{rlc}_i-\eta_Q^1 \nabla_{\theta^{rlc}_i} \mathcal{L}(\phi_i, \theta^{rlc}_i)$
\STATE \hspace{1.0cm} $\phi_i \leftarrow \phi_i-\eta_Q^2 \nabla_{\phi_i} \mathcal{L}(\phi_i, \theta^{rlc}_i)$ for $i=1,2$
\STATE \hspace{1.0cm} $\alpha \leftarrow \alpha-\eta_\alpha \nabla_\alpha \mathcal{L}(\alpha)$
\STATE \hspace{1.0cm} $\bar{\theta}^{rlc}_i \leftarrow \tau^1 \theta^{rlc}_i+(1-\tau^1)\bar{\theta}^{rlc}_i$
\STATE \hspace{1.0cm} $\bar{\phi}_i \leftarrow \tau^2 \phi_i+(1-\tau^2)\bar{\phi}_i$ for $i=1,2$
\STATE \hspace{0.5cm} {\textbf{end for}}
\STATE {\textbf{end for}}
\end{algorithmic}
\label{alg2}
\end{algorithm}

The input of actor network is ${Rep}^{rla}$, and the output is the action distribution $\pi_{\theta}(\cdot \mid {Rep}^{rla})$. MCRL2 employs dual Q-networks to mitigate Q-value overestimation bias. The input of critic network is ${Rep}^{rlc}_i$ and action $a$, output is $Q_{\phi_i}({Rep}^{rlc}_i, a)\ (i=1,2)$ which is used to evaluate the performance of the current policy and directly measure the long-term return of performing action $a$ in state $s$. Besides, we define the target Q-network $Q_{\bar{\phi}_i}$ and target RLC network $\text{RLC}(\cdot; \bar{\theta}^{rlc}_i)$ to stabilize training.

The target Q-value of RL-Critic is defined as follows.

\begin{equation}
\label{eq33}
\begin{aligned}
y =\ & r(s, a) + \gamma \mathbb{E}_{a' \sim \pi_\theta} \bigg[
\min_{i=1,2} \big( Q_{\bar{\phi}_i}(Rep^{rlc'}_i, a') \\
& - \alpha \log \pi_\theta(a' \mid Rep_{rla}') \big)
\bigg]
\end{aligned}
\end{equation}

\noindent where ${Rep}^{rlc'}_i = \text{RLC}(s'; \bar{\theta}^{rlc}_i)$ denotes the learned feature representations of the next state $s'$ using the target RLC network, $a' \sim \pi_\theta (\cdot \mid {Rep}_{rla}')$.

The loss function of RL-Critic is defined as follows.

\begin{equation}
\label{eq34}
\mathcal{L}(\phi_i, \theta^{rlc}_i) = \mathbb{E}_{(s, a, r, s') \sim D} \left[ \left( Q_{\phi_i} ({Rep}^{rlc}_i, a) - y \right)^2 \right]
\end{equation}

\noindent where $i \in \{1, 2\}$.

The policy loss function of RL-Actor is defined as follow.

\begin{equation}
\begin{aligned}
\label{eq35}
\mathcal{L}(\theta, \theta^{rla}) = \mathbb{E}_{s \sim D, a \sim \pi_\theta} \left[ \alpha \log \pi_\theta(a \mid {Rep}^{rla}) \right. \\
\left. - \min_{j=1,2} Q_{\phi_j} ({Rep}^{rlc}_j, a) \right]
\end{aligned}
\end{equation}

The adjustment of temperature $\alpha$ is used to balance reward maximizing and policy entropy. The loss function of $\alpha$ is defined as Eq. \eqref{eq36}.

\begin{equation}
\label{eq36}
\mathcal{L}(\alpha) = \mathbb{E}_{s \sim D, a \sim \pi_\theta} \left[ -\alpha \left( \log \pi_\theta(a \mid {Rep}^{rla}) + \mathcal{H}_{\text{target}} \right) \right]
\end{equation}

\noindent where $\mathcal{H}_{\text{target}} = -|\mathcal{N}|$.

The update of target network is defined as follows.

\begin{equation}
\label{eq37}
\begin{aligned}
\bar{\theta}^{rlc}_i &\leftarrow \tau \theta^{rlc}_i + (1 - \tau) \bar{\theta}^{rlc}_i \\
\bar{\varphi}_i &\leftarrow \tau \varphi_i + (1 - \tau) \bar{\varphi}_i
\end{aligned}
\end{equation}

To summarize, the learning process of MCRL2 is formulated in Algorithm \ref{alg:alg2} and the architecture of MCRL2 is shown in Fig. \ref{mcrl2}. 

\subsection{Convergence Proof of MCRL2}

\noindent {\bf{Lemma 1: The convergence of soft policy evaluation of RL-Critic.}} 

\noindent {\bf{Statement:}} For the fixed policy $\pi$ and the given stable representations ${Rep}^{rla}$ and ${Rep}^{rlc}$, the RL-soft bellman operator $\mathcal{T}^{\pi}$ is defined by Eq. \eqref{eq38}.

\begin{equation}
\label{eq38}
\begin{aligned}
\mathcal{T}^\pi Q({Rep}^{rlc}(s), a) = r(s, a) 
+ \gamma \, \mathbb{E}_{s' \sim P,\, a' \sim \pi} \big[ \\
\quad Q({Rep}^{rlc}(s'), a') 
- \alpha \log \pi(a' \mid {Rep}^{rla}(s')) \big]
\end{aligned}
\end{equation}

Q function sequence $\{Q_k\}$ converge to the only fixed point $Q^{\pi}$.

\begin{equation}
\label{eq39}
\lim_{k \to \infty} Q_k = Q^\pi
\end{equation}

\noindent {\bf{Proof:}} Take any state s and action a, let its corresponding representations be ${Rep}^{rla}(s)$ and ${Rep}^{rlc}(s)$. We have

\begin{equation}
\label{eq40}
\begin{aligned}
&\left| \mathcal{T}^\pi Q_1({Rep}^{rlc}(s), a) - \mathcal{T}^\pi Q_2({Rep}^{rlc}(s), a) \right| \\
=\; & \Big| r(s,a) + \gamma \, \mathbb{E}_{s' \sim P,\, a' \sim \pi} [ Q_1({Rep}^{rlc}(s'), a')  \\
&\quad - \alpha \log \pi(a' \mid {Rep}^{rla}(s')) ] \\
&\quad - r(s,a) - \gamma \, \mathbb{E}_{s' \sim P,\, a' \sim \pi} [ Q_2({Rep}^{rlc}(s'), a') \\
&\quad + \alpha \log \pi(a' \mid {Rep}^{rla}(s'))] \Big| \\
=\; & \gamma \left| \mathbb{E}_{s' \sim P,\, a' \sim \pi} \left[ Q_1({Rep}^{rlc}(s'), a') - Q_2({Rep}^{rlc}(s'), a') \right] \right| \\
\leq\; & \gamma \, \mathbb{E}_{s' \sim P,\, a' \sim \pi} \left| Q_1({Rep}^{rlc}(s'), a') - Q_2({Rep}^{rlc}(s'), a') \right| \\
\leq\; & \gamma \, \lVert Q_1 - Q_2 \rVert_\infty
\end{aligned}
\end{equation}

By taking the upper bound of the above inequality, we can obtain

\begin{equation}
\label{eq41}
\left\| \mathcal{T}^\pi Q_1 - \mathcal{T}^\pi Q_2 \right\|_\infty \leq \gamma \left\| Q_1 - Q_2 \right\|_\infty
\end{equation}

\noindent where $\gamma \in [0,1)$. Thus, $\mathcal{T} ^ \pi$ is a $\gamma$-contraction mapping, for any $Q_1$ and $Q_2$. According to the Banach fixed point theorem, for any initial Q function $Q_0$, we define iteration as follows.

\begin{equation}
\label{eq42}
Q_{k+1} = \mathcal{T}^\pi Q_k
\end{equation}

There exists a unique fixed point $Q^\pi$, and we obtain the desired result Eq. \eqref{eq39}.


$Q_k$ converges to the unique fixed point $Q^\pi$. $\hfill \square$

\noindent {\bf{Lemma 2: The monotonicity of RL-soft policy improvement.}} 

\noindent {\bf{Statement:}} Let the current policy be $\pi_k$, and the corresponding soft Q-function be $Q^{\pi_k}(\mathrm{Rep}^{rlc}(s), a)$ (according to Lemma 1), RL-Actor updates policy by the objective function as Eq. \eqref{eq44}.

\begin{equation}
\label{eq44}
\begin{aligned}
\mathcal{J}(\pi) =\ & \mathbb{E}_{(s_t,a_t) \sim \rho_\pi} \bigg[ \sum_{t=0}^{\infty} \gamma^t \big( r({Rep}_t^{rla}(s), a_t) \\
& + \alpha \mathcal{H}(\pi(\cdot \mid {Rep}_t^{rla}(s))) \big) \bigg]
\end{aligned}
\end{equation}

According to the Bellman equation definition, when $Q^(\pi_k)$ converges, the above equation is equivalent to Eq. \eqref{eq45}.

\begin{equation}
\label{eq45}
\begin{aligned}
\mathcal{J}(\pi) =\ & \mathbb{E}_{s \sim D} \bigg[ \mathbb{E}_{a \sim \pi} \big( Q^{\pi_k}({Rep}_t^{rla}(s), a) \\
& + \alpha \mathcal{H}(\pi(\cdot \mid {Rep}_t^{rla}(s))) \big) \bigg]
\end{aligned}
\end{equation}

\noindent where $\mathcal{H}(\pi(\cdot \mid \mathrm{Rep}_t^{rla}(s))) = -\mathbb{E}_{a \sim \pi} [\log \pi(a \mid \mathrm{Rep}_t^{rla}(s))]$.

Let

\begin{equation}
\label{eq46}
\pi_{k+1} = \arg\max_{\pi} \mathcal{J}(\pi)
\end{equation}

After updating through RL-soft policy improvement, for any state $s$, there are

\begin{equation}
\label{eq47}
\mathcal{J}(\pi_{k+1}) \geq \mathcal{J}(\pi_k)
\end{equation}

For any state $s$, there is

\begin{equation}
\label{eq48}
Q^{\pi_{k+1}}({Rep}^{rla}(s), \pi_{k+1}(s)) \geq Q^{\pi_k}({Rep}^{rla}(s), \pi_k(s))
\end{equation}

\noindent {\bf{Proof:}} Given any state $s$, record its representations ${Rep}^{rla}(s)$. Define a single step objective function.

\begin{equation}
\label{eq49}
\begin{aligned}
\mathcal{J}_s(\pi) =\ & \mathbb{E}_{a \sim \pi} \big[ Q^{\pi_k}({Rep}^{rla}(s), a) \\
& - \alpha \log \pi(a \mid {Rep}^{rla}(s)) \big]
\end{aligned}
\end{equation}

The overall policy objective function is

\begin{equation}
\label{eq50}
\mathcal{J}(\pi) = \mathbb{E}_{s \sim D} \left[ J_s(\pi) \right]
\end{equation}

We introduce the soft policy improvement formula. The target distribution is defined as Eq. \eqref{eq51}. which is a Boltzmann distribution derived from the soft Q-function under the current policy. During policy improvement, $\pi_{k+1}$ is typically updated by minimizing the KL divergence to the target distribution.

\begin{equation}
\label{eq51}
p(s \mid a) \triangleq \frac{\exp\left(\frac{1}{\alpha} Q^{\pi_k}({Rep}^{rla}(s), a)\right)}{N({Rep}^{rla}(s))}
\end{equation}

where the normalization constant is

\begin{equation}
\label{eq52}
{N}({Rep}^{rla}(s)) = \int \exp\left(\frac{1}{\alpha} Q^{\pi_k}({Rep}^{rla}(s), a)\right) \, da
\end{equation}

The single step objective can be rewritten as

\begin{equation}
\label{eq53}
\begin{aligned}
\mathcal{J}_s (\pi) &= \mathbb{E}_{a \sim \pi} [ \alpha ( \frac{1}{\alpha} Q^{\pi_k}({Rep}^{rla}(s), a) \\
& \quad - \log \pi(a \mid {Rep}^{rla}(s)) ) ] \\
&= \alpha \mathbb{E}_{a \sim \pi} [ \log \frac{\exp\left(\frac{1}{\alpha} Q^{\pi_k}({Rep}^{rla}(s), a)\right)}{\pi(a \mid {Rep}^{rla}(s))} ] \\
&= -\alpha \, D_{KL} \left( \pi(\cdot \mid {Rep}^{rla}(s)) \parallel p(s \mid a) \right) \\
& \quad +  \alpha \log Z({Rep}^{rla}(s))
\end{aligned}
\end{equation}

Clearly, $D_{KL}$ is non-negative, with equality if and only if

\begin{equation}
\label{eq54}
\pi(\cdot \mid {Rep}^{rla}(s)) = p(s \mid a)
\end{equation}

where $D_{KL}$ achieves its minimum value of zero. This ensures that $\mathcal{J}_s (\pi)$ is maximized, implying that the optimal policy in each state must satisfy

\begin{equation}
\label{eq55}
\pi_{k+1}(a \mid {Rep}^{rla}(s)) = \frac{\exp\left(\frac{1}{\alpha} Q^{(\pi_k)}({Rep}^{rla}(s), a)\right)}{N({Rep}^{rla}(s))}
\end{equation}

Given any state $s$, we can obtain Eq. \eqref{eq47}. The expectation of the overall objective function is expressed as

\begin{equation}
\label{eq56}
\mathcal{J}(\pi_{k+1}) = \mathbb{E}_{s \sim D} [\mathcal{J}_s(\pi_{k+1})] \geq \mathbb{E}_{s \sim D} [\mathcal{J}_s(\pi_k)] = \mathcal{J}(\pi_k)
\end{equation}

By the maximum entropy policy improvement theorem, for any state $s$, we obtain the desired result Eq. \eqref{eq48}. Hence, it is proven that both the overall objective and the value at each state do not decrease after the policy update.  $\hfill \square$

\noindent {\bf{Theorem: Global Convergence of the Policy in MCRL2.}} 

\noindent {\bf{Statement:}} Under the above assumptions including stable representations and finite state and action spaces. Alternating iterations of RL-soft policy evaluation and RL-soft policy improvement, the policy sequence ${\pi_k}$ converges to the optimal policy $\pi^*$ under the maximum entropy objective.

\noindent {\bf{Proof:}} According to Lemma 1, for a fixed policy $pi_k$, the RL-critic uses the RL-soft Bellman operator $\mathcal{T}^{\pi_k}$. The iteration Eq. \eqref{eq42} is a $\gamma$-contraction. Therefore, by the Banach fixed point theorem, it converges to the unique fixed point $Q^{\pi_k}$.

According to Lemma 2, under the representations ${Rep}^{rla}(s)$, by using the RL-soft policy improvement equation Eq. \eqref{eq55}, the overall objective function meets Eq. \eqref{eq47}.

In each iteration:

\begin{itemize}

\item{For the given policy $\pi_k$, it computes $Q^{\pi_k}$ by RL-soft policy evaluation, which converges uniquely.}

\item{Then, using the RL-soft policy improvement update, the policy is updated to $\pi_{k+1}$ and $\mathcal{J}(\pi_{k+1}) \geq \mathcal{J}(\pi_k)$.}

\end{itemize}

Since the cumulative reward and entropy regularization are bounded in finite state and action spaces, the objective function $\mathcal{J}(\pi)$ is upper bounded, and the sequence $\{\mathcal{J}(\pi_k)\}$ is monotonically non-decreasing and bounded, meaning it converges to a limit.

\begin{equation}
\label{eq57}
\lim_{k \to \infty} J(\pi_k) = J(\pi^*)
\end{equation}

Meanwhile, by the RL-soft policy improvement theorem, for each state $s$, we have Eq. \eqref{eq48}.

Consequently, the policy sequence ${\pi_k}$ converges to the optimal policy $\pi^*$ under the maximum entropy RL objective, which is defined in Eq. \eqref{eq32}. $\hfill \square$

\section{Performance Evaluation}
\label{sec-Eval}

We use Python 3.9.0 and PyTorch 1.10.0 to implement MCRL2. Based on the implementation, the performance evaluation are conducted on a workstation with Intel(R) Core(TM) i7-11800H 2.30GHz CPU, 32GB memory and a NVIDIA GeForce RTX 3080 GPU. This section presents experiments conducted to address the following research questions. Code availability: https://github.com/igeng/MSSchedRL.

\begin{itemize}

\item{\emph{RQ1:} How effective is MCRL2 in training convergence compared with DRL-based baselines?}

\item{\emph{RQ2:} How effective is MCRL2 in load balancing?}

\item{\emph{RQ3:} How effective is MCRL2 in microservice scheduling success rate?}

\item{\emph{RQ4:} How effective is MCRL2 in average completion time of instance?}

\item{\emph{RQ5:} How efficient is MCRL2 in scheduling decision time?}

\end{itemize}

\subsection{Experimental Setup}

{\bf{Cluster Configuration.}} We simulate a cluster consisting of 15 heterogeneous nodes. As illustrated in Table \ref{tab:table2}, three types of VM instances are selected, each with distinct resource configurations.

\begin{table}[!t]
\caption{CLUSTER CONFIGURATION DETAILS\label{tab:table2}}
\centering
\begin{tabular}{c c c c}
\hline
Types & CPU Cores & Memory Capacity & VM Quantity\\
\hline
1 & 32 & 32 & 5\\
2 & 32 & 64 & 5\\
3 & 64 & 128 & 5\\
\hline
\end{tabular}
\end{table}

{\bf{Microservice Workload Patterns.}} The experiments are conducted based on Alibaba cluster traces microservices v2021, which are collected from Alibaba production clusters. The traces record the CPU and memory utilization of over 90,000 containers belonging to more than 1,300 microservices in the same production cluster over 12 hours. Specifically, we select 2,000 microservice instance workload records, ordered by arrival time, as our training dataset. For evaluation, we employ two datasets extracted sequentially from the same data source: one containing the next 2,000 microservice instances and another containing the next 4,000 microservice instances. All microservice instance durations are modeled as following a Poisson distribution with $\lambda=600s$. Finally, we separately construct two microservice workload patterns, MS-2000 and MS-4000.

{\bf{Baselines.}} We conducted a series of experiments comparing MCRL2 against the following baselines. 

\begin{itemize}
\item{\emph{Round Robin.}
Microservice instances are allocated to nodes in a cyclic and uniform manner.}

\item{\emph{Random algorithm.}
Microservice instances are assigned to nodes in a stochastic fashion.}

\item{\emph{Double Deep Q-Network (DDQN).}
DDQN combines Q-learning with deep neural networks and decouples action selection and evaluation for more stable policy learning in complex environments \cite{ref6} \cite{ref7}.}

\item{\emph{Proximal Policy Optimization (PPO).} 
PPO employs an actor-critic architecture and stabilizes policy updates through a clipped objective function \cite{ref5} \cite{ref8}.}

\item{\emph{Soft Actor-Critic.}
SAC enhances exploration by maximizing policy entropy, making it particularly effective for dynamic environments \cite{ref9} \cite{ref10}.}
\end{itemize}

{\bf{Ablation study setting.}} In our implementation, SAC employs standard multilayer perceptrons for both actor and critic networks, following common practice in prior work. This design not only serves as a fair baseline but also provides the control architecture for our ablation studies, where the MLP modules are replaced with MCRL-based representations to assess the effectiveness of the proposed representation learning.

{\bf{Evaluation metrics.}} The following three metrics are used as the basis of evaluating the comparison algorithms.

(1) DCLB (degree of cumulative load balancing)

DCLB represents the aggregated load balancing performance across all scheduling steps. A lower DCLB value reflects a more effective load balancing performance of the algorithm. The load balancing degree of the $j$-th resource dimension, denoted as $\mathcal{LB}j$, is defined in Eq. \eqref{eq9}. Accordingly, the load balancing degree for the $j$-th resource dimension at scheduling step $t$ is represented as $\mathcal{LB}{t}^{j}$. The DCLB for the $j$-th resource dimension is therefore formulated as follows.

\begin{align}
\label{dclb_j}
    {DCLB}^{j} &= \sum_{t=1}^{T} \mathcal{LB}_{t}^{j}
\end{align}

\noindent where $T$ indicates the total scheduling steps. In the experiments, the resource dimensions considered are CPU and memory.

(2) DRLB (degree of real-time load balancing)

DRLB reflects the load balancing level of the cluster at a given scheduling step. The cumulative sum of DRLB values across all scheduling steps corresponds to the DCLB. In contrast to DCLB, temporal variations in DRLB reveal the dynamic load balancing performance of the approach across different scheduling steps. Following each scheduling step, the DRLB values of all approaches are ranked in ascending order. The cumulative frequency of each algorithm at each rank is recorded, where a higher frequency at top ranks indicates superior load balancing performance.

(3) MSSR (microservice scheduling success rate)

If the constraint defined in Eq. \eqref{eq6} is satisfied, the microservice instance is considered successfully scheduled. The MSSR is defined in Eq. \eqref{eq11}. An increase in MSSR indicates an improvement in the QoS of microservice systems.

(4) ACTI (average completion time of instance)

The average completion time of instance is used to further evaluate the performance of the approach and is defined in Eq. \eqref{acti}.

\begin{equation}
\label{acti}
    ACTI = \frac {Total\ completion\ time} {\sum_{k=1}^{\lvert \mathcal{MS} \rvert} \lvert \mathcal{MI}_k \rvert}
\end{equation}

\subsection{RQ1: Effectiveness for training convergence.}

The evaluation metric employed is the average episodic reward achieved throughout the training process. In particular, we study the trend of the episode increases. A faster convergence, characterized by a more rapid stabilization of the reward curve, indicates higher learning efficiency. Furthermore, a higher stabilized reward value after convergence reflects superior overall performance of the proposed approach.

\begin{figure}[!t]
\centering
\includegraphics[width=0.4\textwidth]{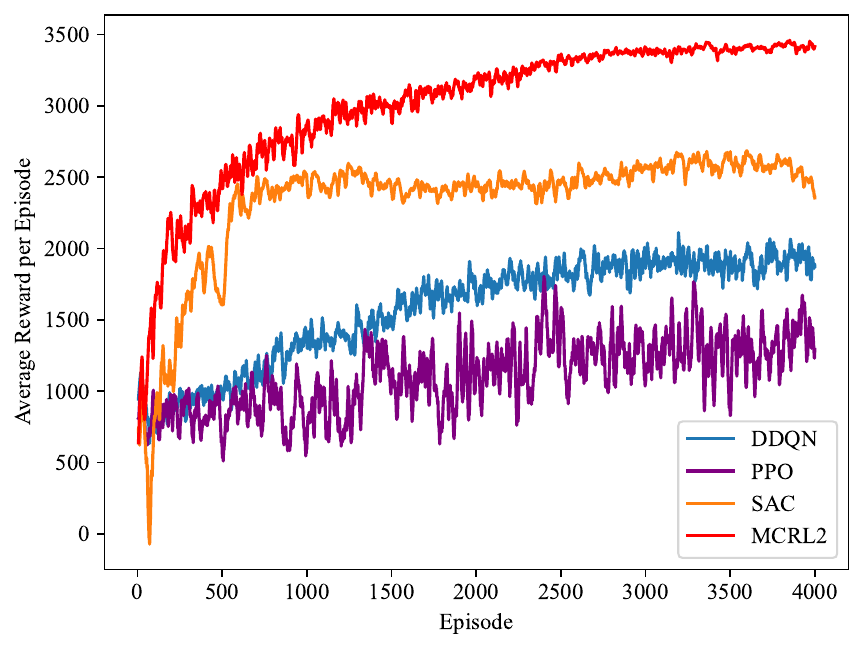}
\caption{Comparison of Average Episode Rewards Across DRL-based Algorithms.}
\label{RL_conver}
\end{figure}

As illustrated in Fig. \ref{RL_conver}, MCRL2 achieves the fastest convergence and the highest converged reward compared to the baselines. During the initial training phase, SAC demonstrates a convergence speed comparable to that of MCRL2. However, its final converged reward is notably lower. This indicates that MCRL effectively extracts key information from the original state, thereby accelerating convergence, improving data efficiency, and enhancing the overall convergence performance of RL. Moreover, the approaches based on DDQN and PPO exhibit not only slower convergence but also lower converged reward. In particular, the PPO-based approach exhibits significant reward fluctuations during training. This instability can be attributed to its reliance on high-variance on-policy sampling and low sample efficiency. Additionally, the clipping mechanism, which restricts the magnitude of policy updates, may cause under-optimization, further contributing to instability during the training process.

\subsection{RQ2: Effectiveness for load balancing.}

The second experiment focuses on evaluating the load balancing performance of MCRL2 by comparing it with five baselines: Round Robin, Random, DDQN, PPO, and SAC.

\begin{figure}[!t]
\centering
\includegraphics[width=0.4\textwidth]{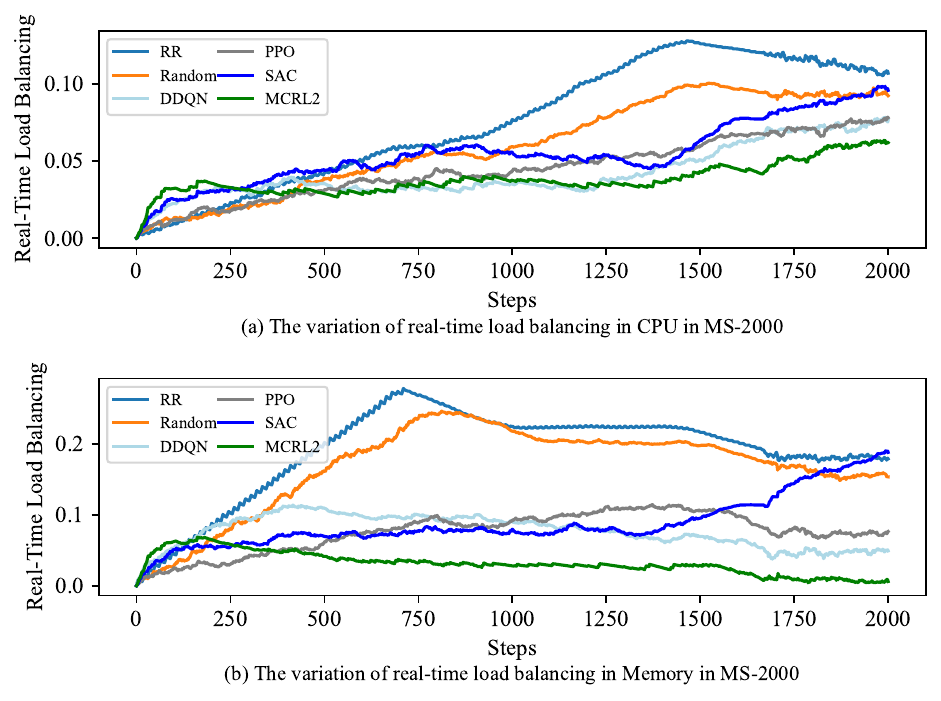}
\caption{Comparison of Real-Time Load Balancing Over Time Under MS-2000 Pattern.}
\label{rtlb-2000}
\end{figure}

\begin{figure}[!t]
\centering
\includegraphics[width=0.4\textwidth]{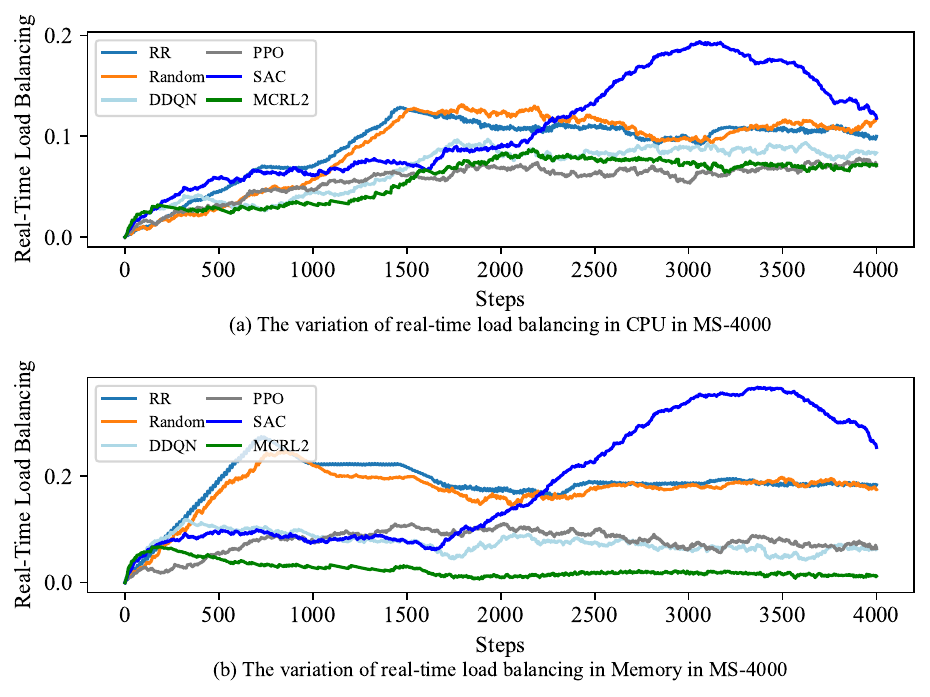}
\caption{Comparison of Real-Time Load Balancing Over Time Under MS-4000 Pattern.}
\label{rtlb-4000}
\end{figure}

{\bf{Real-Time Load Balancing Over Time.}} We first evaluate the real-time load balancing degree of all approaches at each timestep in CPU and memory dimensions under the MS-2000 and MS-4000 patterns. The metric is defined such that lower values correspond to better load balancing performance, dynamically reflecting the effectiveness of scheduling policies. As shown in Figs. \ref{rtlb-2000} and \ref{rtlb-4000}, MCRL2 consistently maintains lower levels of imbalance in both CPU and memory across different time stages, indicating more effective microservice scheduling. During the initial scheduling phase, all approaches exhibit comparable performance. However, as illustrated in Figs. \ref{rtlb-2000}(b) and \ref{rtlb-4000}(b), the rule-based approaches rapidly demonstrate degraded performance under both patterns, highlighting the greater challenge of memory-aware scheduling. In contrast, MCRL2 achieves the best memory load balancing around the 500th timestep and sustains this advantage throughout the end of scheduling. Under the more challenging MS-4000 pattern, SAC performs worst in the latter half, which contrasts with its good reward convergence during training. This discrepancy suggests that, although SAC benefits from efficient exploration and fast convergence in the training, its inherent policy stochasticity and overfitting to training distribution result in insufficient generalization in test environments with distributional shifts and dynamic workloads.  

\begin{figure}[!t]
\centering
\includegraphics[width=0.4\textwidth]{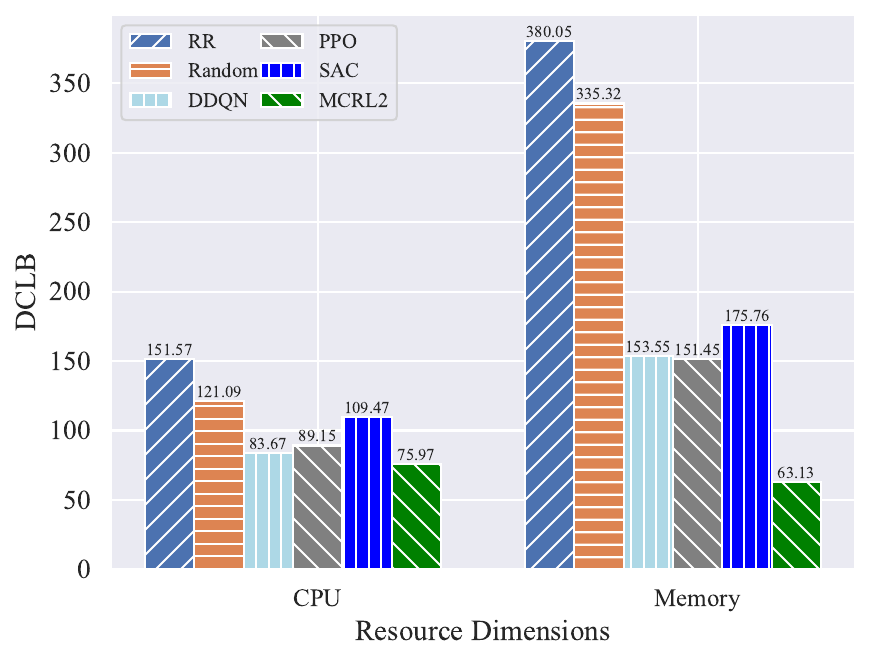}
\caption{Comparison of Cumulative Load Balancing Under MS-2000 Pattern.}
\label{dclb-2000}
\end{figure}

\begin{figure}[!t]
\centering
\includegraphics[width=0.4\textwidth]{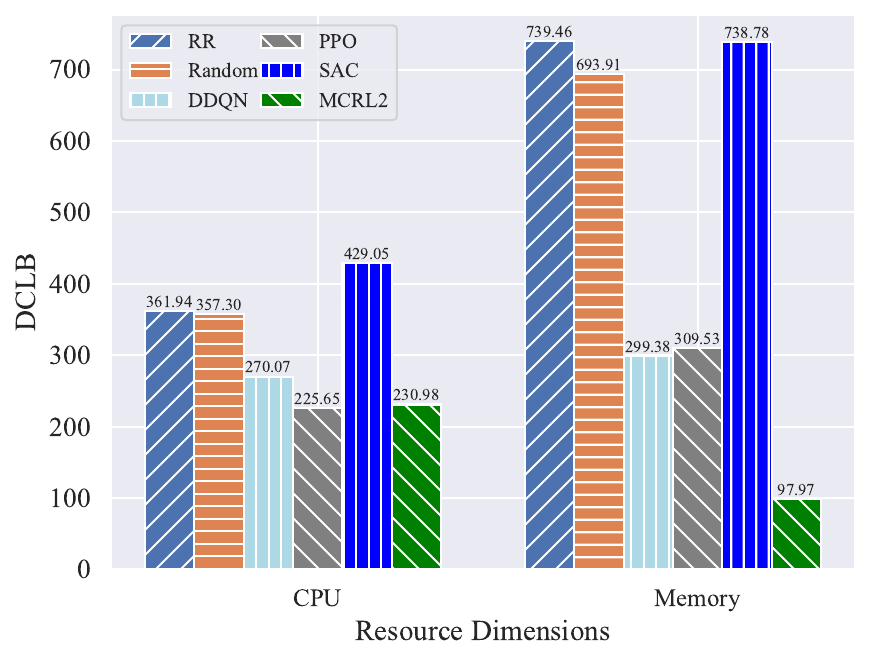}
\caption{Comparison of Cumulative Load Balancing Under MS-4000 Pattern.}
\label{dclb-4000}
\end{figure}

{\bf{Cumulative Load Balancing.}} In addition to real-time performance, we quantitatively assess the overall load balancing performance by accumulating the degree of load balancing across the evaluation period. Figs. \ref{dclb-2000} and \ref{dclb-4000} illustrate the degree of cumulative load balancing across different resource dimensions and patterns. MCRL2 achieves comprehensive superiority. Notably, as shown in Figs. \ref{dclb-2000}(b) and \ref{dclb-4000}(b), it significantly outperformed baselines in the memory dimension.

\begin{figure}[!t]
\centering
\includegraphics[width=0.4\textwidth]{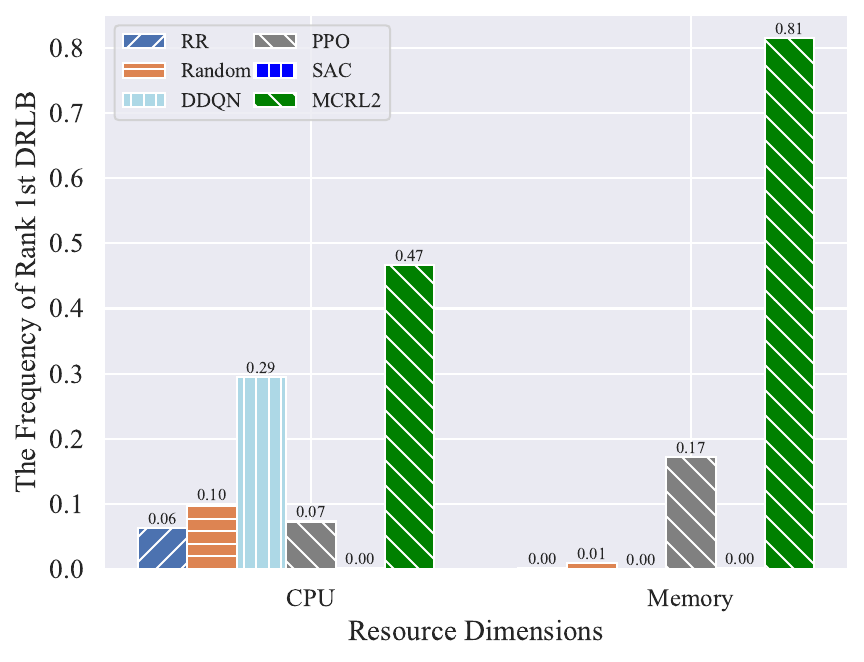}
\caption{Comparison of Frequency of Rank 1st Real-Time Load Balancing Under MS-2000 Pattern.}
\label{drlb-2000}
\end{figure}

\begin{figure}[!t]
\centering
\includegraphics[width=0.4\textwidth]{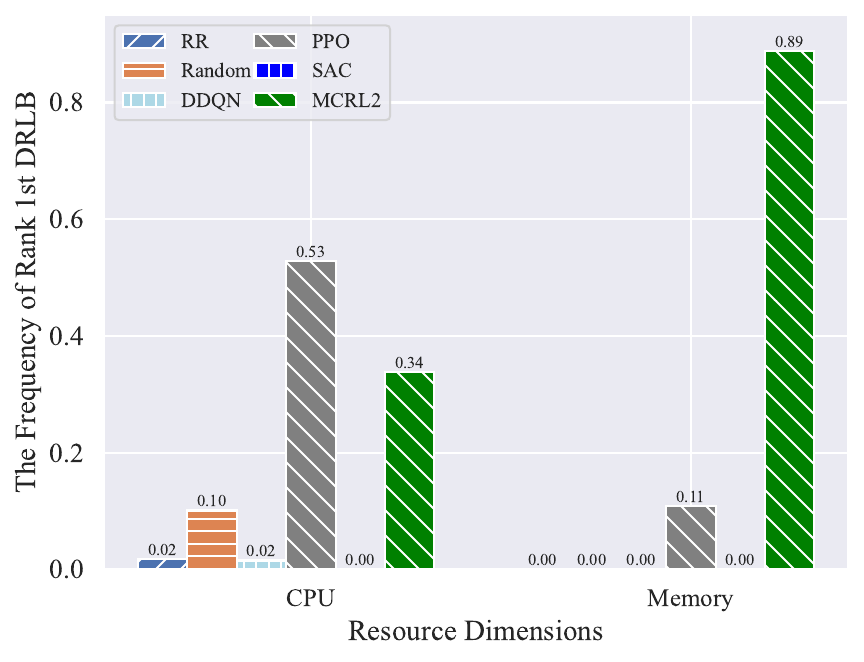}
\caption{Comparison of Frequency of Rank 1st Real-Time Load Balancing Under MS-4000 Pattern.}
\label{drlb-4000}
\end{figure}

{\bf{Frequency of Rank 1st Real-Time Load Balancing.}} Finally, we statistically analyzes the frequency with which each approach achieves the lowest real-time load balancing (i.e., optimal performance) in CPU and memory dimensions. This further supports the superiority of MCRL2. As shown in Figs. \ref{drlb-2000} and \ref{drlb-4000}, MCRL2 consistently ranks first more frequently than the baselines. When considering the CPU results from Figs. \ref{dclb-4000} and \ref{drlb-4000}, MCRL2 only performs slightly worse than PPO. However, the difference is minor and the cumulative CPU load imbalance under the MS-4000 patterns differs by only 5.33. In all other cases, MCRL2 significantly outperforms PPO.

Considering that the two experimental patterns encompass microservice instances with diverse duration and resource demands from the real production clusters, the results conclusively demonstrate that MCRL2 achieves significantly better load balancing performance than all baselines. Moreover, through representation learning in MCRL, MCRL2 exhibits superior generalization across different workload patterns. These findings indicate that MCRL not only accelerates convergence and improves final rewards during training but also significantly enhances load balancing performance under different microservice patterns.

\subsection{RQ3: Effectiveness for microservice scheduling success rate.}

We compare the scheduling success rate of MCRL2 with five baseline approaches. The scheduling success rate is defined as the proportion of microservice instances successfully scheduled without any resource conflicts, as specified in Eq. \eqref{eq11}.

\begin{figure}[!t]
\centering
\includegraphics[width=0.4\textwidth]{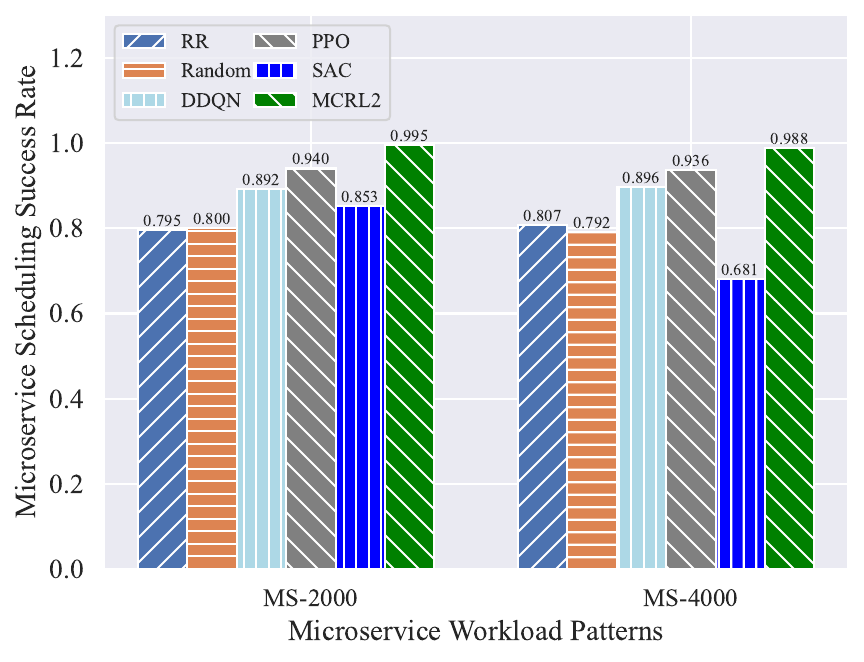}
\caption{Comparison of Microservice Scheduling Success Rate.}
\label{mssr}
\end{figure}

As illustrated in Fig. \ref{mssr}, MCRL2 achieves the highest scheduling success rate under MS-2000 and MS-4000 patterns. This superior performance demonstrates MCRL2 can effectively handle inherent resource constraints in cloud environments, ensuring a higher proportion of successful scheduling. Specifically, MCRL2 consistently maintains the best performance, achieving success rates of 0.995 and 0.988, respectively. These results represent improvements of 5.85\% and 5.56\% over the best baseline approach in each pattern, indicating a significant advantage.

Further analysis reveals that the observed improvement in success rate is primarily attributed to MCRL. By leveraging multi-resource cross-attention-based representation learning at both the cluster and node levels, the approach can more effectively capture characteristics of resource capacities across the cluster. When resource capacities vary significantly across dimensions and nodes, MCRL2 demonstrates higher scheduling efficiency. It assigns microservice instances with diverse resource demands to the most appropriate nodes, thereby minimizing resource contention.

\subsection{RQ4: Effectiveness for average completion time.}

As shown in Fig. \ref{acti}, MCRL2 achieves the lowest average completion time of instance in both evaluation patterns. The reduction in completion time not only reflects efficient resource utilization but also contributes to improving overall system throughput. MCRL2 reduces the average completion time to 603.33 seconds  and 607.28 seconds, representing improvements of 5.48\% and 5.29\%, respectively, compared to the best baseline. More importantly, beyond reducing average completion time, MCRL2 achieves superior load balancing and higher scheduling success rates. This is more challenging than merely ensuring successful scheduling, as simply selecting the node with the most resources would leads to load imbalance.

\begin{figure}[!t]
\centering
\includegraphics[width=0.4\textwidth]{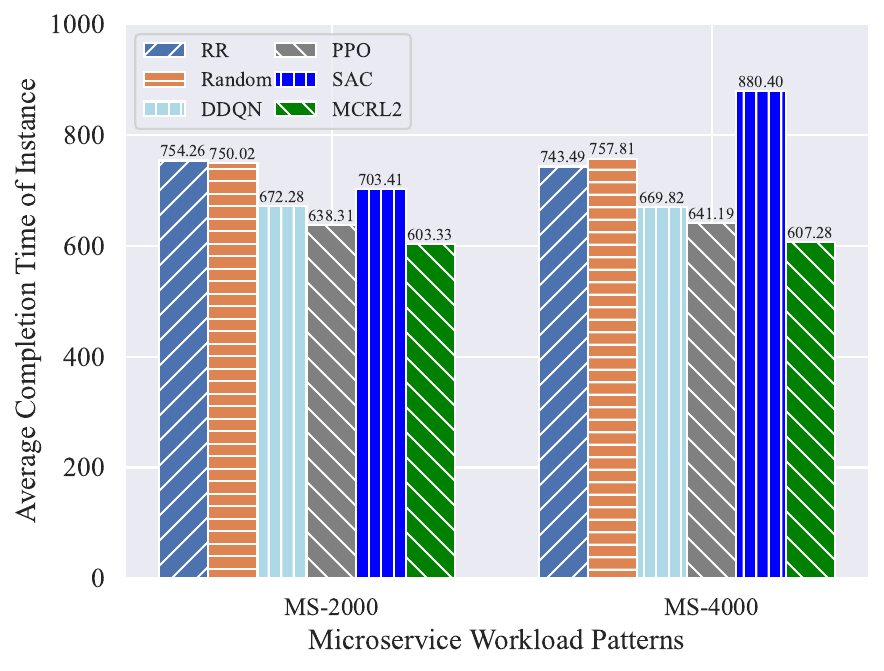}
\caption{Comparison of Average Completion Time of Instance.}
\label{acti}
\end{figure}

\subsection{RQ5: Efficiency for scheduling decision time.}

\begin{table}[!t]
\caption{Cumulative scheduling decision time of DRL-based approaches under different microservice workload patterns (in seconds).\label{tab:table_time}}
\centering
\begin{tabular}{p{0.8in}<{\centering}p{1in}<{\centering}p{1in}<{\centering}}
\hline
Algorithms & MS-2000 & MS-4000 \\
\hline
DDQN & 1.441 & 1.977 \\
PPO & 1.786 & 2.703 \\
SAC & 1.856 & 3.163 \\
MCRL2 & 2.645 & 4.686 \\
\hline
\end{tabular}
\end{table}

As shown in Table \ref{tab:table_time}, the reported values represent the cumulative scheduling decision time under each workload pattern, averaged over five experimental runs. Although MCRL2 incurs higher cumulative time compared to the DRL-based baselines, the additional delay remains relatively small. More importantly, considering the significant improvements achieved by MCRL2 in other evaluation metrics, the extra inference time does not constitute a performance bottleneck.

\section{Conclusion}

In this paper, we propose MCRL2, a novel reinforcement learning approach enhanced with multi-resource cross-attention-based representation learning, to tackle key challenges in microservice scheduling, including dynamic resource imbalance, nonlinear cross-resource coupling, and multi-dimensional request heterogeneity. By capturing fine-grained interdependencies among nodes, resources, and microservices, the proposed MCRL enables informative state representations that significantly improve scheduling effectiveness. Integrated with soft actor-critic, MCRL2 achieves stable and efficient decision-making. Extensive experiments on real-world production traces demonstrate that MCRL2 consistently outperforms baselines in terms of load balancing, scheduling success rate, and average completion time of instance, highlighting its practical value in dynamic microservice environments.

\bibliographystyle{IEEEtran}
\bibliography{references.bib}

@INPROCEEDINGS{ref1,
  author={Santos, José and Wauters, Tim and Turck, Filip De and Steenkiste, Peter},
  booktitle={2024 33rd International Conference on Computer Communications and Networks (ICCCN)}, 
  title={Towards Optimal Load Balancing in Multi-Zone Kubernetes Clusters via Reinforcement Learning}, 
  year={2024},
  volume={},
  number={},
  pages={1-9},
  doi={10.1109/ICCCN61486.2024.10637606}}

@ARTICLE{ref2,
  author={Lv, Wenkai and Wang, Quan and Yang, Pengfei and Ding, Yunqing and Yi, Bijie and Wang, Zhenyi and Lin, Chengmin},
  journal={IEEE Transactions on Parallel and Distributed Systems}, 
  title={Microservice Deployment in Edge Computing Based on Deep Q Learning}, 
  year={2022},
  volume={33},
  number={11},
  pages={2968-2978},
  doi={10.1109/TPDS.2022.3150311}}

@article{ref3,
  title={DRS: A deep reinforcement learning enhanced Kubernetes scheduler for microservice-based system},
  author={Jian, Zhaolong and Xie, Xueshuo and Fang, Yaozheng and Jiang, Yibing and Lu, Ye and Dash, Ankan and Li, Tao and Wang, Guiling},
  journal={Software: Practice and Experience},
  volume={54},
  number={10},
  pages={2102--2126},
  year={2024},
  publisher={Wiley Online Library}
}

@INPROCEEDINGS{ref4,
  author={Gu, Lin and Zeng, Deze and Hu, Jie and Li, Bo and Jin, Hai},
  booktitle={IEEE INFOCOM 2021 - IEEE Conference on Computer Communications}, 
  title={Layer Aware Microservice Placement and Request Scheduling at the Edge}, 
  year={2021},
  volume={},
  number={},
  pages={1-9},
  doi={10.1109/INFOCOM42981.2021.9488779}}

@ARTICLE{ref5,
  author={Wang, Liangyuan and Liu, Xudong and Ding, Haonan and Hu, Yi and Peng, Kai and Hu, Menglan},
  journal={IEEE Transactions on Computers}, 
  title={Energy-Delay-Aware Joint Microservice Deployment and Request Routing With DVFS in Edge: A Reinforcement Learning Approach}, 
  year={2025},
  volume={74},
  number={5},
  pages={1589-1604},
  doi={10.1109/TC.2025.3535826}}

@article{ref6,
title = {Microservice instances selection and load balancing in fog computing using deep reinforcement learning approach},
journal = {Future Generation Computer Systems},
volume = {156},
pages = {77-94},
year = {2024},
issn = {0167-739X},
doi = {https://doi.org/10.1016/j.future.2024.03.010},
url = {https://www.sciencedirect.com/science/article/pii/S0167739X24000815},
author = {Wassim Boudieb and Abdelhamid Malki and Mimoun Malki and Ahmed Badawy and Mahmoud Barhamgi}}

@article{ref7,
title = {Multi-user edge service orchestration based on Deep Reinforcement Learning},
journal = {Computer Communications},
volume = {203},
pages = {30-47},
year = {2023},
issn = {0140-3664},
doi = {https://doi.org/10.1016/j.comcom.2023.02.027},
url = {https://www.sciencedirect.com/science/article/pii/S0140366423000737},
author = {Christian Quadri and Alberto Ceselli and Gian Paolo Rossi}
}

@article{ref8,
  title={Proximal policy optimization algorithms},
  author={Schulman, John and Wolski, Filip and Dhariwal, Prafulla and Radford, Alec and Klimov, Oleg},
  journal={arXiv preprint arXiv:1707.06347},
  year={2017}
}

@InProceedings{ref9,
  title = 	 {Soft Actor-Critic: Off-Policy Maximum Entropy Deep Reinforcement Learning with a Stochastic Actor},
  author =       {Haarnoja, Tuomas and Zhou, Aurick and Abbeel, Pieter and Levine, Sergey},
  booktitle = 	 {Proceedings of the 35th International Conference on Machine Learning},
  pages = 	 {1861--1870},
  year = 	 {2018},
  editor = 	 {Dy, Jennifer and Krause, Andreas},
  volume = 	 {80},
  series = 	 {Proceedings of Machine Learning Research},
  month = 	 {10--15 Jul},
  publisher =    {PMLR},
  url = 	 {https://proceedings.mlr.press/v80/haarnoja18b.html}
}

@ARTICLE{ref10,
  author={Wang, Jinming and Li, Shaobo and Zhang, Xingxing and Zhu, Keyu and Xie, Cankun and Wu, Fengbin},
  journal={IEEE Access}, 
  title={Deep Reinforcement Learning Task Scheduling Method for Real-Time Performance Awareness}, 
  year={2025},
  volume={13},
  number={},
  pages={31385-31400},
  doi={10.1109/ACCESS.2025.3534980}}

@article{ref11, title={Scheduling of Time-Varying Workloads Using Reinforcement Learning}, volume={35}, url={https://ojs.aaai.org/index.php/AAAI/article/view/17088}, DOI={10.1609/aaai.v35i10.17088}, abstractNote={Resource usage of production workloads running on shared compute clusters often fluctuate significantly across time. While simultaneous spike in the resource usage between two workloads running on the same machine can create performance degradation, unused resources in a machine results in wastage and undesirable operational characteristics for a compute cluster. Prior works did not consider such temporal resource fluctuations or their alignment for scheduling decisions. Due to the variety of time-varying workloads, their complex resource usage characteristics, it is challenging to design well-defined heuristics for scheduling them optimally across different machines in a cluster. In this paper, we propose a Deep Reinforcement Learning (DRL) based approach to exploit various temporal resource usage patterns of time varying workloads as well as a technique for creating equivalence classes among a large number of production workloads to improve scalability of our method. Validations with real production traces from Google and Alibaba show that our technique can significantly improve metrics for operational excellence (e.g. utilization, fragmentation, resource exhaustion etc.) for a cluster, compared to the baselines.}, number={10}, journal={Proceedings of the AAAI Conference on Artificial Intelligence}, author={Mondal, Shanka Subhra and Sheoran, Nikhil and Mitra, Subrata}, year={2021}, month={May}, pages={9000-9008} }

@inproceedings{ref12,
author = {Luo, Shutian and Xu, Huanle and Lu, Chengzhi and Ye, Kejiang and Xu, Guoyao and Zhang, Liping and Ding, Yu and He, Jian and Xu, Chengzhong},
title = {Characterizing Microservice Dependency and Performance: Alibaba Trace Analysis},
year = {2021},
isbn = {9781450386388},
publisher = {Association for Computing Machinery},
address = {New York, NY, USA},
url = {https://doi.org/10.1145/3472883.3487003},
doi = {10.1145/3472883.3487003},
booktitle = {Proceedings of the ACM Symposium on Cloud Computing},
pages = {412–426},
numpages = {15},
location = {Seattle, WA, USA},
series = {SoCC '21}
}

@ARTICLE{ref13,
  author={Yu, Xiaoming and Wu, Wenjun and Wang, Yangzhou},
  journal={IEEE Transactions on Services Computing}, 
  title={Integrating Cognition Cost With Reliability QoS for Dynamic Workflow Scheduling Using Reinforcement Learning}, 
  year={2023},
  volume={16},
  number={4},
  pages={2713-2726},
  doi={10.1109/TSC.2023.3253182}}

@article{ref14,
  title={Deep reinforcement learning-based methods for resource scheduling in cloud computing: A review and future directions},
  author={Zhou, Guangyao and Tian, Wenhong and Buyya, Rajkumar and Xue, Ruini and Song, Liang},
  journal={Artificial Intelligence Review},
  volume={57},
  number={5},
  pages={124},
  year={2024},
  publisher={Springer}
}

@ARTICLE{ref15,
  author={Zhong, Chenxing and Li, Shanshan and Huang, Huang and Liu, Xiaodong and Chen, Zhikun and Zhang, Yi and Zhang, He},
  journal={IEEE Transactions on Software Engineering}, 
  title={Domain-Driven Design for Microservices: An Evidence-Based Investigation}, 
  year={2024},
  volume={50},
  number={6},
  pages={1425-1449},
  doi={10.1109/TSE.2024.3385835}}

@ARTICLE{ref16,
  author={Luo, Shutian and Xu, Huanle and Lu, Chengzhi and Ye, Kejiang and Xu, Guoyao and Zhang, Liping and He, Jian and Xu, Chengzhong},
  journal={IEEE Transactions on Parallel and Distributed Systems}, 
  title={An In-Depth Study of Microservice Call Graph and Runtime Performance}, 
  year={2022},
  volume={33},
  number={12},
  pages={3901-3914},
  doi={10.1109/TPDS.2022.3174631}}

@INPROCEEDINGS{ref17,
  author={Lin, Chenyu and Luo, Shutian and Xu, Huanle},
  booktitle={2023 IEEE Intl Conf on Parallel \& Distributed Processing with Applications, Big Data \& Cloud Computing, Sustainable Computing \& Communications, Social Computing \& Networking (ISPA/BDCloud/SocialCom/SustainCom)}, 
  title={Exploring Imbalances among Microservice Containers in Large Cloud Platforms}, 
  year={2023},
  volume={},
  number={},
  pages={237-245},
  doi={10.1109/ISPA-BDCloud-SocialCom-SustainCom59178.2023.00064}}

@ARTICLE{ref18,
  author={He, Xiang and Tu, Zhiying and Wagner, Markus and Xu, Xiaofei and Wang, Zhongjie},
  journal={IEEE Transactions on Cloud Computing}, 
  title={Online Deployment Algorithms for Microservice Systems With Complex Dependencies}, 
  year={2023},
  volume={11},
  number={2},
  pages={1746-1763},
  doi={10.1109/TCC.2022.3161684}}

@ARTICLE{ref19,
  author={Zhu, Jianyong and Yang, Renyu and Sun, Xiaoyang and Wo, Tianyu and Hu, Chunming and Peng, Hao and Xiao, Junqing and Zomaya, Albert Y. and Xu, Jie},
  journal={IEEE Transactions on Parallel and Distributed Systems}, 
  title={QoS-Aware Co-Scheduling for Distributed Long-Running Applications on Shared Clusters}, 
  year={2022},
  volume={33},
  number={12},
  pages={4818-4834},
  doi={10.1109/TPDS.2022.3202493}}

@ARTICLE{ref20,
  author={Shi, Jiuchen and Fu, Kaihua and Wang, Jiawen and Chen, Quan and Zeng, Deze and Guo, Minyi},
  journal={IEEE Transactions on Parallel and Distributed Systems}, 
  title={Adaptive QoS-Aware Microservice Deployment With Excessive Loads via Intra- and Inter-Datacenter Scheduling}, 
  year={2024},
  volume={35},
  number={9},
  pages={1565-1582},
  doi={10.1109/TPDS.2024.3425931}}

@ARTICLE{ref21,
  author={He, Xiang and Xu, Hanchuan and Xu, Xiaofei and Chen, Yin and Wang, Zhongjie},
  journal={IEEE Transactions on Services Computing}, 
  title={An Efficient Algorithm for Microservice Placement in Cloud-Edge Collaborative Computing Environment}, 
  year={2024},
  volume={17},
  number={5},
  pages={1983-1997},
  doi={10.1109/TSC.2024.3399650}}

@article{ref22,
author = {Pallewatta, Samodha and Kostakos, Vassilis and Buyya, Rajkumar},
title = {Placement of Microservices-based IoT Applications in Fog Computing: A Taxonomy and Future Directions},
year = {2023},
issue_date = {December 2023},
publisher = {Association for Computing Machinery},
address = {New York, NY, USA},
volume = {55},
number = {14s},
issn = {0360-0300},
url = {https://doi.org/10.1145/3592598},
doi = {10.1145/3592598},
journal = {ACM Comput. Surv.},
month = jul,
articleno = {321},
numpages = {43}
}

@ARTICLE{ref23,
  author={Ding, Zhijun and Wang, Song and Jiang, Changjun},
  journal={IEEE Transactions on Cloud Computing}, 
  title={Kubernetes-Oriented Microservice Placement With Dynamic Resource Allocation}, 
  year={2023},
  volume={11},
  number={2},
  pages={1777-1793},
  doi={10.1109/TCC.2022.3161900}}

@article{ref24,
author = {Kumar, Mohit and Samriya, Jitendra Kumar and Dubey, Kalka and Gill, Sukhpal Singh},
title = {QoS-aware resource scheduling using whale optimization algorithm for microservice applications},
journal = {Software: Practice and Experience},
volume = {54},
number = {4},
pages = {546-565},
doi = {https://doi.org/10.1002/spe.3211},
url = {https://onlinelibrary.wiley.com/doi/abs/10.1002/spe.3211},
eprint = {https://onlinelibrary.wiley.com/doi/pdf/10.1002/spe.3211},
year = {2024}
}

@article{ref25,
author = {Xu, Minxian and Yang, Lei and Wang, Yang and Gao, Chengxi and Wen, Linfeng and Xu, Guoyao and Zhang, Liping and Ye, Kejiang and Xu, Chengzhong},
title = {Practice of Alibaba cloud on elastic resource provisioning for large-scale microservices cluster},
journal = {Software: Practice and Experience},
volume = {54},
number = {1},
pages = {39-57},
doi = {https://doi.org/10.1002/spe.3271},
url = {https://onlinelibrary.wiley.com/doi/abs/10.1002/spe.3271},
eprint = {https://onlinelibrary.wiley.com/doi/pdf/10.1002/spe.3271},
year = {2024}
}

@misc{ref26,
      title={Deep Reinforcement Learning for Job Scheduling and Resource Management in Cloud Computing: An Algorithm-Level Review}, 
      author={Yan Gu and Zhaoze Liu and Shuhong Dai and Cong Liu and Ying Wang and Shen Wang and Georgios Theodoropoulos and Long Cheng},
      year={2025},
      eprint={2501.01007},
      archivePrefix={arXiv},
      primaryClass={cs.DC},
      url={https://arxiv.org/abs/2501.01007}, 
}

@article{ref27,
author = {Zhong, Zhiheng and Xu, Minxian and Rodriguez, Maria Alejandra and Xu, Chengzhong and Buyya, Rajkumar},
title = {Machine Learning-based Orchestration of Containers: A Taxonomy and Future Directions},
year = {2022},
issue_date = {January 2022},
publisher = {Association for Computing Machinery},
address = {New York, NY, USA},
volume = {54},
number = {10s},
issn = {0360-0300},
url = {https://doi.org/10.1145/3510415},
doi = {10.1145/3510415},
journal = {ACM Comput. Surv.},
month = sep,
articleno = {217},
numpages = {35}
}

@ARTICLE{ref28,
  author={Liu, Ruochen and Yang, Ping and Lv, Haoyuan and Li, Weibin},
  journal={IEEE Transactions on Cloud Computing}, 
  title={Multi-Objective Multi-Factorial Evolutionary Algorithm for Container Placement}, 
  year={2023},
  volume={11},
  number={2},
  pages={1430-1445},
  doi={10.1109/TCC.2021.3137400}}

@INPROCEEDINGS{ref29,
  author={Zhou, Jiali and Li, Xin and Wang, Qinhui and Qin, Xiaolin and Miao, WeiWei and Tian, Jianwei},
  booktitle={2022 IEEE 28th International Conference on Parallel and Distributed Systems (ICPADS)}, 
  title={Balancing Load: An Adaptive Traffic Management Scheme for Microservices}, 
  year={2023},
  volume={},
  number={},
  pages={641-648},
  doi={10.1109/ICPADS56603.2022.00089}}

@ARTICLE{ref30,
  author={Wei, Wenting and Gu, Huaxi and Wang, Kun and Li, Jianjia and Zhang, Xuan and Wang, Ning},
  journal={IEEE Transactions on Network and Service Management}, 
  title={Multi-Dimensional Resource Allocation in Distributed Data Centers Using Deep Reinforcement Learning}, 
  year={2023},
  volume={20},
  number={2},
  pages={1817-1829},
  doi={10.1109/TNSM.2022.3213575}}

@ARTICLE{ref31,
  author={Hu, Pihe and Chen, Yu and Pan, Ling and Fang, Zhixuan and Xiao, Fu and Huang, Longbo},
  journal={IEEE/ACM Transactions on Networking}, 
  title={Multi-User Delay-Constrained Scheduling With Deep Recurrent Reinforcement Learning}, 
  year={2024},
  volume={32},
  number={3},
  pages={2344-2359},
  doi={10.1109/TNET.2024.3359911}}

@inproceedings{ref32,
 author = {Vaswani, Ashish and Shazeer, Noam and Parmar, Niki and Uszkoreit, Jakob and Jones, Llion and Gomez, Aidan N and Kaiser, \L ukasz and Polosukhin, Illia},
 booktitle = {Advances in Neural Information Processing Systems},
 editor = {I. Guyon and U. Von Luxburg and S. Bengio and H. Wallach and R. Fergus and S. Vishwanathan and R. Garnett},
 pages = {},
 publisher = {Curran Associates, Inc.},
 title = {Attention is All you Need},
 url = {https://proceedings.neurips.cc/paper_files/paper/2017/file/3f5ee243547dee91fbd053c1c4a845aa-Paper.pdf},
 volume = {30},
 year = {2017}
}

@ARTICLE{ref33,
  author={Wang, Ziliang and Zhu, Shiyi and Li, Jianguo and Jiang, Wei and Ramakrishnan, K. K. and Yan, Meng and Zhang, Xiaohong and Liu, Alex X.},
  journal={IEEE/ACM Transactions on Networking}, 
  title={DeepScaling: Autoscaling Microservices With Stable CPU Utilization for Large Scale Production Cloud Systems}, 
  year={2024},
  volume={32},
  number={5},
  pages={3961-3976},
  doi={10.1109/TNET.2024.3400953}}

@inproceedings{ref34,
 author = {Li, Jianing and Papyan, Vardan},
 booktitle = {Advances in Neural Information Processing Systems},
 editor = {A. Oh and T. Naumann and A. Globerson and K. Saenko and M. Hardt and S. Levine},
 pages = {57660--57712},
 publisher = {Curran Associates, Inc.},
 title = {Residual Alignment: Uncovering the Mechanisms of Residual Networks},
 url = {https://proceedings.neurips.cc/paper_files/paper/2023/file/b3f48945f6fb402b4b5cdcf490e72847-Paper-Conference.pdf},
 volume = {36},
 year = {2023}
}

@ARTICLE{ref35,
  author={Yang, Hao and Pan, Li and Liu, Shijun},
  journal={IEEE Internet of Things Journal}, 
  title={An RL-Based Cost-Effective Two-Layer Scaling Strategy for Multiregional Heterogeneous and Time-Varying Cloud Instances}, 
  year={2025},
  volume={12},
  number={8},
  pages={10709-10721},
  doi={10.1109/JIOT.2024.3513958}}

@ARTICLE{ref36,
  author={Bai, Haoyu and Xu, Minxian and Ye, Kejiang and Buyya, Rajkumar and Xu, Chengzhong},
  journal={IEEE Transactions on Services Computing}, 
  title={DRPC: Distributed Reinforcement Learning Approach for Scalable Resource Provisioning in Container-Based Clusters}, 
  year={2024},
  volume={17},
  number={6},
  pages={3473-3484},
  doi={10.1109/TSC.2024.3433388}}

\end{document}